%% file: main_arxiv.tex
\documentclass[10pt,twocolumn,letterpaper]{article}
\PassOptionsToPackage{table}{xcolor}
\usepackage[pagenumbers]{cvpr}
\input{preamble}
\usepackage{tikz}
\input{arxiv_spacing}
\hypersetup{
  pdftitle={Beyond a Single Human Demonstration: Learning from Scored Trajectory Alternatives for End-to-End Driving},
  pdfauthor={Xinyao Rao, Zecheng Tang, Liuxuan Sun, Zheyuan Lai, Minjun Yu, Xun Cao, Wei Li},
  bookmarksnumbered=true
}
\newcommand{\arxivdisplaytitle}{Beyond a Single Human Demonstration:\\
  Learning from Scored Trajectory Alternatives for End-to-End Driving}
\newcommand{\arxivauthors}{%
  Xinyao Rao$^{1}$\hspace{0.65em}%
  Zecheng Tang$^{2}$\hspace{0.65em}%
  Liuxuan Sun$^{1}$\hspace{0.65em}%
  Zheyuan Lai$^{3}$\hspace{0.65em}%
  Minjun Yu$^{3}$\hspace{0.65em}%
  Xun Cao$^{1}$\hspace{0.65em}%
  Wei Li$^{1,3,*}$}
\newcommand{\arxivaffiliations}{$^{1}$Nanjing University\qquad $^{2}$Peking University\qquad $^{3}$EABOT.AI}
\newcommand{\arxivcorrespondence}{$^{*}$Corresponding author.}
\newcommand{\arxivreportdate}{November 20, 2026}
\newcommand{\arxivreportname}{DriveSTA Technical Report}
\definecolor{ArxivHeaderPurple}{HTML}{63065F}
\makeatletter
\newcommand{\arxivheadrule}{{\color{ArxivHeaderPurple}\hrule height .5pt}}
\newcommand{\arxivcoverhead}{%
  \vbox{%
    \hbox to\textwidth{%
      \includegraphics[height=26pt]{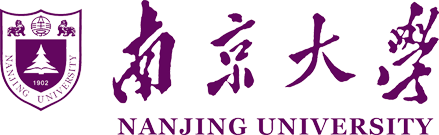}%
      \hspace{16pt}%
      \includegraphics[height=26pt]{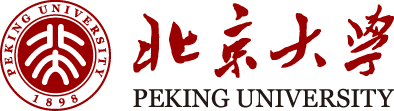}%
      \hspace{16pt}%
      \raisebox{6.75pt}{%
        \includegraphics[height=12.5pt]{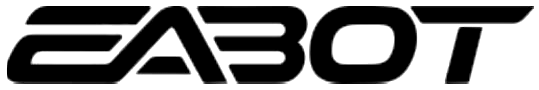}%
      }%
      \hfil
      \raisebox{2pt}{\normalfont\sffamily\bfseries\small\color{ArxivHeaderPurple}\arxivreportdate}%
    }%
    \kern6pt
    \arxivheadrule
  }%
}
\newcommand{\arxivrunninghead}{%
  \vbox{%
    \hbox to\textwidth{%
      {\normalfont\sffamily\bfseries\small\color{ArxivHeaderPurple}\arxivreportname}%
      \hfil
      {\normalfont\sffamily\bfseries\small\color{ArxivHeaderPurple}\thepage}%
    }%
    \kern8pt
    \arxivheadrule
  }%
}
\newcommand{\ps@arxiv}{%
  \ps@plain
  \let\@oddhead\arxivrunninghead
  \let\@evenhead\arxivrunninghead
  \let\@oddfoot\@empty
  \let\@evenfoot\@empty
}
\newcommand{\ps@arxivcover}{%
  \ps@plain
  \let\@oddhead\arxivcoverhead
  \let\@evenhead\arxivcoverhead
  \def\@oddfoot{\hfil{\normalfont\color{ArxivHeaderPurple}\thepage}\hfil}%
  \let\@evenfoot\@oddfoot
}
\makeatother
\title{\arxivdisplaytitle}
\author{\arxivauthors\\\arxivaffiliations\\{\small\arxivcorrespondence}}
\date{}
\begin{document}
\pagestyle{arxiv}
\onecolumn
\thispagestyle{arxivcover}
\input{sections/00_arxiv_frontmatter}
\twocolumn
\begingroup
\mainbodylayout

\arxivtextspacing
\arxivmainflow
\input{sections/01_introduction}
\input{sections/02_overview_figure}
\input{sections/02_related_work}
\input{sections/03_method}
\arxivfloatspacing
\input{sections/04_experiments}
\input{sections/05_discussion}
\input{sections/06_conclusion}
\par
{\setlength{\parskip}{0pt}\small\input{sections/08_references}}
\clearpage
\endgroup

\appendix
\setcounter{table}{0}\setcounter{equation}{0}\setcounter{figure}{0}
\renewcommand{\thetable}{S\arabic{table}}
\renewcommand{\theequation}{S\arabic{equation}}
\renewcommand{\thefigure}{S\arabic{figure}}
\renewcommand{\theHtable}{supp.\arabic{table}}
\renewcommand{\theHequation}{supp.\arabic{equation}}
\renewcommand{\theHfigure}{supp.\arabic{figure}}
\onecolumn
\arxivtextspacing
\begin{center}
\Large\textbf{\papertitle}\\[0.5em]
Supplementary Material
\end{center}
\setlength{\multicolsep}{6pt}
\captionsetup{skip=4pt,justification=justified,singlelinecheck=false}
\begin{multicols}{2}
\suppressfloats[t]
\newcommand{\suppTableStyle}{\fontsize{8.5}{10}\selectfont\setlength{\tabcolsep}{3pt}\renewcommand{\arraystretch}{1.13}}
Appendix~\ref{app:datasets} specifies the evaluation protocols. Appendix~\ref{app:details} collects implementation and training details. Appendix~\ref{app:qualitative} presents NAVSIM comparisons and HUGSIM video sequences.

\input{sections/10_appendix_datasets_metrics}
\input{sections/11_appendix_experimental_details}

\input{sections/14_appendix_hugsim_breakdown}
\end{multicols}
\input{sections/14_appendix_hugsim_table}
\clearpage
\input{sections/13_appendix_qualitative}
\end{document}

%% file: preamble.tex
\usepackage{amsmath,amssymb,mathtools}
\usepackage{graphicx,booktabs,multirow,array}
\usepackage{xcolor}
\definecolor{TableHeader}{HTML}{E9EFF5}
\definecolor{TableGroup}{HTML}{F0F3F6}
\definecolor{TableAlt}{HTML}{F7F9FB}
\usepackage{microtype,placeins}
\usepackage[colaction]{multicol}
\definecolor{cvprblue}{rgb}{0.21,0.49,0.74}
\definecolor{TableOurs}{HTML}{EBF5FC}
\usepackage[pagebackref,breaklinks,colorlinks,allcolors=cvprblue]{hyperref}
\usepackage[capitalize]{cleveref}
\newcommand{\method}{DriveSTA}
\newcommand{\papertitle}{Beyond a Single Human Demonstration:\\Learning from Scored Trajectory Alternatives for End-to-End Driving}
\newcommand{\sg}{\operatorname{sg}}

\newcommand{\Jhat}{\widehat J}

\DeclareMathOperator*{\argmin}{arg\,min}

\newcommand{\ScaledRefinedBest}{95.55}

\newcommand{\FinalBest}{95.14}

\makeatletter
\newcommand{\mainbodylayout}{%
  \ifnum\c@page=1\relax
    \@tempdima=\dbltextfloatsep
    \advance\@tempdima by -6pt
    \ifdim\@tempdima>0pt
      \global\advance\@colht by \@tempdima
      \global\advance\@colroom by \@tempdima
      \global\vsize=\@colroom
    \fi
  \fi
  \raggedbottom
  \def\@textbottom{\ifnum\c@page=2\relax
    \if@firstcolumn\else\vskip\z@ plus .0001fil\fi
    \else\vskip\z@ plus .0001fil\fi}%
  \clubpenalty=3000
  \widowpenalty=3000
  \displaywidowpenalty=3000
  \brokenpenalty=1000
  \setlength{\parskip}{0pt}%
  \setlength{\textfloatsep}{4pt}%
  \setlength{\dbltextfloatsep}{6pt}%
  \setlength{\floatsep}{3pt}%
  \setlength{\dblfloatsep}{4pt}%
  \setlength{\intextsep}{4pt}%
  \captionsetup{skip=4pt}%
  \setlength{\abovedisplayskip}{3pt}%
  \setlength{\belowdisplayskip}{3pt}%
  \setlength{\abovedisplayshortskip}{3pt}%
  \setlength{\belowdisplayshortskip}{3pt}%
  \def\cvprsection{\@startsection{section}{1}{\z@}%
    {-6pt plus -2pt minus -2pt}{4pt}{\large\bf}}%
  \def\cvprsubsection{\@startsection{subsection}{2}{\z@}%
    {-5pt plus -2pt minus -2pt}{3pt}{\elvbf}}%
  \renewcommand{\paragraph}{\@startsection{paragraph}{4}{\z@}%
    {0pt}{-1em}{\normalfont\normalsize\bfseries}}%
}
\makeatother

%% file: sections/00_arxiv_frontmatter.tex
\begingroup
\setlength{\parindent}{0pt}
\setlength{\parskip}{0pt}
\colorlet{ArxivCard}{ArxivHeaderPurple!8!white}
\definecolor{ArxivLink}{HTML}{365C70}
\hypersetup{urlcolor=ArxivLink}
\input{sections/00_arxiv_icons}
\vspace*{-2.5pt}
\noindent
\begin{tikzpicture}
  \node[fill=ArxivCard,draw=none,rounded corners=9pt,
        inner xsep=14pt,inner ysep=12pt,outer sep=0pt] {%
    \begin{minipage}{\dimexpr\textwidth-28pt\relax}
      \begin{center}
        {\fontsize{15.5}{18.5}\selectfont\bfseries
        \arxivdisplaytitle\par}
        \vspace{8pt}
        {\fontsize{10.3}{12.5}\selectfont
          \arxivauthors\par}
        \vspace{3pt}
        {\fontsize{9.5}{11.5}\selectfont
          \arxivaffiliations\par}
      \end{center}
      \vspace{1pt}
      {\fontsize{10}{12}\selectfont\normalfont
        \textbf{Abstract.}\enspace
        \input{sections/00_abstract}
      }
      \vspace{8pt}
      {\fontsize{9.3}{11.5}\selectfont\urlstyle{same}
        \begin{tabular}{@{}c@{\hspace{0.55em}}l@{\hspace{0.7em}}l@{}}
          \arxivmailicon & \textbf{Email:} & \href{mailto:yoloredal@gmail.com}{yoloredal@gmail.com}\\
          \arxivcodeicon & \textbf{Code:} & \url{https://github.com/Rtwotwo/DriveSTA}\\
          \arxivprojecticon & \textbf{Project:} & \url{https://rtwotwo.github.io/DriveSTA}\\
          \arxivhficon & \textbf{Hugging Face:} & \url{https://huggingface.co/Rtwotwo/DriveSTA}
        \end{tabular}\par
      }
      \vspace{4pt}
      {\fontsize{8.7}{10.5}\selectfont \arxivcorrespondence\par}
    \end{minipage}%
  };
\end{tikzpicture}\par
\vspace{10pt}
\begingroup
\setkeys{Gin}{trim={6bp 0bp 6.5bp 0bp},clip}
\input{sections/00_teaser}
\endgroup
\endgroup

%% file: sections/01_introduction.tex
\section{Introduction}

End-to-end autonomous driving aims to learn safe and effective driving behavior from sensor observations and human experience. Recent work has combined autoregressive trajectory generation and diffusion-based modeling in two-stage planners that first propose candidate trajectories and then refine them. ChainFlow-VLA~\cite{wang2026chainflow} follows this direction, using an autoregressive generator to produce multiple proposals and a proposal-conditioned residual flow model to refine each trajectory as a whole.

The effectiveness of this proposal--refinement framework depends on the supervision available for both stages. Although ChainFlow-VLA generates multiple candidates, each driving log records only one human demonstration. Its winner-take-all objective assigns the closest proposal to that reference, leaving other feasible driving choices without explicit trajectory targets. The demonstration alone also provides limited information about the relative quality of these alternatives. To incorporate scene semantics into refinement, ChainFlow-VLA conditions its residual flow model on an online vision-language model (VLM). This semantic guidance improves performance in ambiguous and long-tail scenarios, but running the VLM adds substantial inference cost. Furthermore, refinement in interactive traffic requires context about how surrounding agents move over time. This calls for temporal representations that complement the scene semantics provided by VLMs.

We introduce \method{}, which retains ChainFlow-VLA's probabilistic formulation and uses scored alternatives to guide both trajectory generation and residual denoising. For each training scene, we search over smooth deformations of the human trajectory and score each candidate by simulating its execution against logged traffic. We match diverse, high-scoring alternatives to distinct autoregressive proposals. For refinement, each proposal can be paired with several compatible alternatives, each defining a residual from the proposal to a trajectory with high execution quality. These residuals provide the targets for quality-guided residual denoising. Candidates spanning the full score range, including failed trajectories, train the scorer to assess driving quality. Because refinement can change a candidate's quality, we evaluate refined trajectories again during training and use their updated scores to supervise selection.

To provide temporal context for both refinement and selection, \method{} uses a joint-embedding predictive architecture (JEPA) adapted through masked video-feature prediction~\cite{wang2026drivejepa}. At inference, a shared encoder processes the observed frames once. The denoiser uses the resulting temporal representation to refine each proposal, and the scorer uses the same representation to assess the refined trajectories. This design provides scene context for refinement while avoiding the cost of a separate large model for online semantic conditioning.

On NAVSIM v1 Navtest, \method{} ranks first among the compared planners trained on NAVSIM data alone. With expanded training data, \method{}-Scale achieves state-of-the-art results among the compared methods on both Navtest and closed-loop HUGSIM, reaching 95.6 PDMS on Navtest. In matched evaluations, \method{}-Scale reduces mean planner forward-pass latency by 73\% relative to ChainFlow-VLA. Our contributions are threefold.

\begin{itemize}
    \item We construct scored trajectory alternatives through offline search and simulated execution, providing diverse supervision beyond the single human demonstration. Across 95,998 real scenes with retained targets, our search yields 2.61M screened trajectories.
    \item We extend the proposal-conditioned residual flow framework with quality-guided denoising and integrate a shared temporal representation for refinement and rescoring.
    \item We demonstrate strong planning performance on NAVSIM and closed-loop HUGSIM, with lower forward-pass latency than ChainFlow-VLA under matched conditions.
\looseness=-1\relax\par
\end{itemize}

%% file: sections/02_overview_figure.tex
\begin{figure*}[t]
\centering
\captionsetup{skip=3pt,justification=justified,singlelinecheck=false}\includegraphics[width=\textwidth]{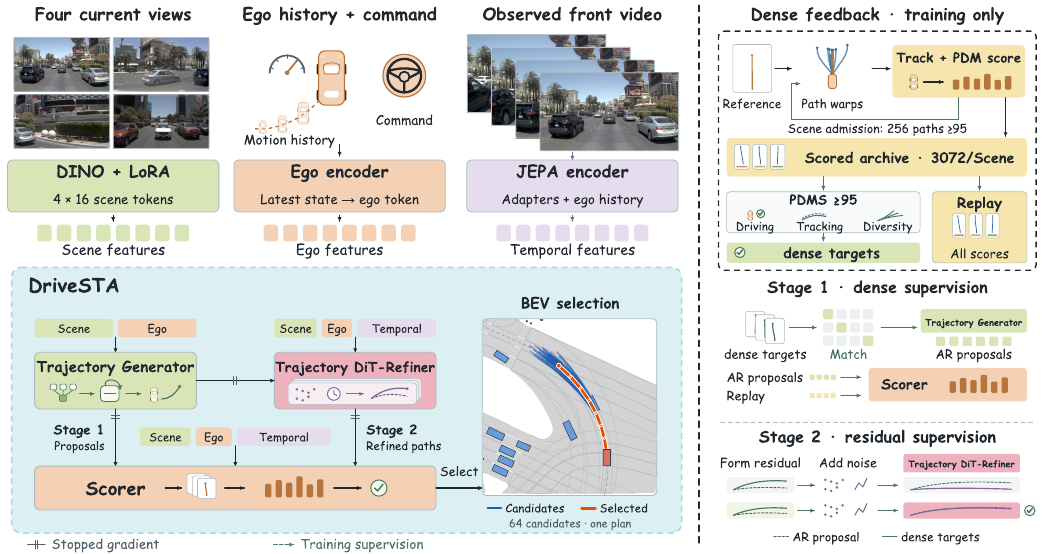}
\caption{\textbf{DriveSTA overview.} Screened targets supervise generation and residual refinement. Evaluator labels train scoring, with fresh labels after refinement. Shared observed-frame JEPA features condition both modules. Double bars mark detached coordinates.}
\label{fig:overview}
\end{figure*}

%% file: sections/02_related_work.tex
\section{Related Work}
\paragraph{Multimodal planning.}
Driving scenes require candidate coverage and reliable selection. DiffusionDrive~\cite{diffusiondrive2024} denoises clustered anchors, whereas ChainFlow-VLA~\cite{wang2026chainflow} refines autoregressive proposals through VLM-conditioned residual diffusion. Both use demonstration-based geometric supervision (\cref{fig:teaser}(a,b)). GTRS~\cite{gtrs2025} scores dense vocabularies and generated proposals, while ZTRS~\cite{ztrs2025} learns a discrete policy from rule-based rewards. These quality signals support selection. DriveSTA adds execution-validated geometric targets for generation and residual refinement in ChainFlow-VLA.
\looseness=-1\relax\par

\paragraph{Trajectory supervision.}
Alternative supervision depends on target coverage and validity. Drive-JEPA~\cite{wang2026drivejepa} filters a clustered vocabulary through simulation, while CLOVER~\cite{clover2026} constructs maneuver-based pseudo-experts for generator self-distillation. Their initial coverage depends on the vocabulary or maneuver set. FPD-Drive~\cite{chen2026fpd} distills simulator-filtered alternatives from a teacher with recorded future observations to a student using current observations. DriveZero~\cite{drivezero2026} distills privileged reinforcement-learning rollouts into a camera-based student. Teacher coverage therefore bounds these targets. DriveSTA searches and screens trajectories within each logged scene (\cref{fig:teaser}(c)), using screened alternatives for geometric supervision and successful and failed executions for scoring.
\looseness=-1\relax\par

\paragraph{Video representations.}
Trajectory assessment requires context about surrounding traffic motion. Predictive training shapes representations through masked video learning in Drive-JEPA~\cite{wang2026drivejepa} and future-feature supervision in FLARE~\cite{flare2026}, which further applies reinforcement fine-tuning. Other approaches make predicted futures part of planning. Auto-JEPA~\cite{autojepa2026} retrieves trajectories using predicted intent, DriveFuture~\cite{drivefuture2026} conditions diffusion on future latents, and WA-JEPA~\cite{wajepa2026} jointly predicts scene latents and actions. To provide temporal context without online VLM inference, DriveSTA adapts a Drive-JEPA encoder and shares its observed-frame features across refinement and scoring. Recorded future frames support feature-prediction training, whereas refinement and scoring use only observed-frame encoder outputs during both training and inference.
\looseness=-1\relax\par

%% file: sections/03_method.tex
\section{Method}
\label{sec:method}
\subsection{Problem formulation and overview}
Each logged scene $s$ provides one reference trajectory $e_s$, leaving alternative trajectories without explicit targets or execution scores. Given four current camera views, observed front-camera frames, ego-state history, and a navigation command, DriveSTA selects one of $N=64$ candidate trajectories. Each $\tau\in\mathbb R^{H\times3}$ comprises $H=8$ ego-frame poses $(x,y,\theta)$ at 0.5-second intervals.
\looseness=-1\relax\par

To supply the missing supervision, we construct and evaluate alternatives offline within each scene (\cref{fig:overview}). Each tracked trajectory receives six component labels $y_s(\tau)\in[0,1]^6$ and an aggregate score $J_s(\tau)\in[0,1]$, with predictive driver model score (PDMS) equal to $100J$. Screened targets supervise geometry, whereas scores from successful and failed executions supervise candidate quality. Refined candidates receive fresh evaluator labels, and refinement and scoring share one JEPA encoding of observed frames.
\looseness=-1\relax\par

\subsection{Constructing scored trajectory supervision}
\label{sec:construction}
\noindent\textbf{Trajectory search.} We seek variations in path and speed anchored at the reference's first sampled position. We therefore apply smooth longitudinal and lateral deformations. Following CLOVER~\cite{clover2026}, we map an ego-fixed offset $u\in\mathbb R^2$ to the position at sample $t$ through a quintic smoothstep
\looseness=-1\relax\par
\begin{equation}
\begin{aligned}
 x_t(u)&=e_t^{xy}+b(r_t)u,\quad r_t=\frac{t-1}{H-1},\\
 b(r)&=6r^5-15r^4+10r^3.
\end{aligned}
\label{eq:warp}
\end{equation}
Here $e_t^{xy}$ denotes the reference position. The deformation preserves the first sampled position and reaches the requested endpoint offset, with zero displacement slope at both ends. Fixed sampling times allow positional changes to vary speed. Headings follow path tangents, retaining reference headings on nearly stationary segments. We track each complete trajectory and evaluate its execution against logged traffic.
\looseness=-1\relax\par

To preserve broad coverage while concentrating queries around promising solutions, each 64-query batch allocates 25\% to broad exploration and the remainder to neighborhoods of high-scoring offsets. When additional search is required, we expand the region and increase the query budget from 3,072 to 8,192 and then 32,768. Score learning requires both successes and failures. We therefore retain 3,072 trajectories across score ranges in $B_s$, independently of the query count. Scene admission requires 256 distinct paths with $J\geq0.95$, independently of the screening budget. Search bounds and recovery budgets are specified in the supplement.
\looseness=-1\relax\par

\noindent\textbf{Target screening.} High aggregate scores alone do not ensure trackability or diversity. We construct the target set $A_s$ by requiring full scores for no-at-fault collision (NC), drivable-area compliance (DAC), time-to-collision (TTC), comfort (C), and driving-direction compliance (DDC), alongside checks on direction, vehicle footprint, motion continuity, and tracking error. Thresholds are specified in the supplement.
\looseness=-1\relax\par

We visit archived trajectories in descending score order and retain those with $J\geq0.95$ that pass all checks. To avoid redundant supervision, a candidate must differ from every retained target by at least 0.25\,m in whole-path average displacement error (ADE). Screening stops at 32 targets or 256 tracking-validation queries. Stage~1 rotates up to eight targets per visit, whereas Stage~2 uses the full screened set for local matching. Scenes without screened targets retain expert supervision, and failed trajectories remain in $B_s$ for score learning.
\looseness=-1\relax\par

\subsection{Stage 1: Learning from scored alternatives}
\label{sec:stage1}
\noindent\textbf{Diverse proposal supervision.} Expert anchoring $\mathcal L_E$ supervises only the proposal closest to each reference through mean L1 pose error. To cover additional feasible choices, we assign screened targets to distinct proposals. In Stage~1, DINOv2~\cite{oquab2024dinov2} with low-rank adaptation (LoRA)~\cite{hu2022lora} encodes the four views into scene tokens $S_s$. Following ChainFlow-VLA~\cite{wang2026chainflow}, an autoregressive generator combines these tokens with the current ego state to produce proposals $P_s=\{p_i\}_{i=1}^{N}$ through recurrent controls and bicycle integration. At epoch $e$, Hungarian matching assigns rotating targets $A_s^{(e)}=\{h_j\}_{j=1}^{K_s}$ to distinct proposals.
\looseness=-1\relax\par
\begin{equation}
\begin{aligned}
 \pi^*&=\argmin_{\pi\ \mathrm{injective}}\sum_{j=1}^{K_s}d(p_{\pi(j)},h_j),\\
 \mathcal L_A&=\frac1{K_s}\sum_{j=1}^{K_s}d(p_{\pi^*(j)},h_j).
\end{aligned}
\label{eq:assignment}
\end{equation}
Here $\pi$ maps target indices to distinct proposal indices. The cost $d$ averages the sum of absolute $x,y$ errors and 0.25 times absolute heading error over the trajectory. Unlike expert anchoring, heading differences are wrapped to $[-\pi,\pi]$. Separated targets and injective assignment encourage proposals to cover distinct alternatives. Matching indices are fixed during backpropagation, and empty target sets contribute zero. Where available, a five-second reference provides a time-compressed auxiliary target for the eight prediction slots, as detailed in the supplement. Unavailable synthetic long-reference supervision contributes zero. The generator objective is $\mathcal L_G=\mathcal L_E+0.25r(e)\mathcal L_A$, where $r(e)=\min(1,e/2)$ introduces dense supervision over the first two epochs.
\looseness=-1\relax\par

\noindent\textbf{Execution score supervision.} Geometric coverage alone does not identify which candidate will execute successfully. The scorer embeds candidate coordinates with a multilayer perceptron (MLP), attends to scene tokens, and incorporates an ego-state token. Six sigmoid heads predict the five components above and ego progress (EP). Following the v1 evaluator, $\Jhat$ multiplies predicted NC and DAC by the weighted mean of EP, TTC, and comfort, with weights 5, 5, and 2. DDC receives supervision without entering the aggregate. To supervise component predictions alongside overall quality, calibration combines component-wise binary cross-entropy (BCE) with aggregate-score error over a valid candidate set $V$.
\looseness=-1\relax\par
\begingroup\small\thinmuskip=1mu\medmuskip=2mu\thickmuskip=2mu\def\mintagsep{2pt}
\begin{equation}
 \mathcal L_{\rm cal}(V)=\frac1{|V|}\sum_{i\in V}
 \left[\sum_{c=1}^6\mathrm{BCE}(\hat y_{ic},y_{ic})+
 0.5(\Jhat_i-J_i)^2\right].
\label{eq:cal}
\end{equation}
\endgroup
All 64 native candidates receive fresh evaluator labels. To preserve exposure to failures as the generator evolves, we replay 64 archived paths across high, medium, low-positive, and zero scores. The scoring objective is $\mathcal L_S(C_s)=\mathcal L_{\rm cal}(C_s)+0.25\mathcal L_{\rm cal}(V_s^{\rm replay})$, with $C_s=P_s$ in Stage~1. Replay uses a separate attention context to leave native predictions unchanged. Padding is masked, empty sets contribute zero, and evaluator soft labels are preserved.
\looseness=-1\relax\par

\subsection{Stage 2: Multi-target residual learning}
\label{sec:stage2}
\noindent\textbf{Multi-target residual supervision.} ChainFlow-VLA~\cite{wang2026chainflow} supervises residual refinement using the recorded trajectory, leaving other feasible choices without refinement targets. Its denoising loss is restricted to the proposal nearest to the demonstration. We extend its diffusion transformer (DiT) to learn residual corrections from multiple execution-validated alternatives constructed in Section~\ref{sec:construction}. Since alternatives may encode different maneuvers, we restrict supervision to geometrically compatible targets to keep corrections local. Each detached proposal $\bar p_i=\sg(p_i)$ is associated with up to four such targets from $A_s$. Positional ADE, endpoint distance, and mean absolute wrapped heading error must not exceed 3\,m, 6\,m, and 0.5\,rad. Position distances are Euclidean. We retain the nearest qualifying targets according to $d$, with indices $\mathcal N_i$.
\looseness=-1\relax\par

Each target $h_j$ defines a residual $\delta_{ij}=h_j\ominus\bar p_i$, where $\ominus$ wraps heading differences. Thus, one proposal can receive several candidate corrections. We sample one diffusion timestep per scene and one Gaussian noise tensor $\epsilon_i$ per proposal, shared across its targets. DiT learns to reconstruct these residuals from noise using $v$-prediction~\cite{salimans2022progressive}, conditioned on the parent, timestep, current ego motion, and scene/video tokens.
\looseness=-1\relax\par
\begin{equation}
\begin{aligned}
 z_{ij,t}&=\alpha_t\delta_{ij}+\sigma_t\epsilon_i,\\
 v_{ij,t}&=\alpha_t\epsilon_i-\sigma_t\delta_{ij},\\
 \hat\tau_{ij}&=\bar p_i\oplus(\alpha_tz_{ij,t}-\sigma_t\hat v_{ij,t}),
\end{aligned}
\label{eq:reconstruct}
\end{equation}
Here $\alpha_t$ is the square root of the cumulative signal schedule, $\sigma_t=\sqrt{1-\alpha_t^2}$, and $\oplus$ wraps reconstructed headings. The predicted clean residual is added to its parent. To learn a specific alternative without averaging distinct corrections, we select the target with the smallest reconstruction error and use that same pair for both losses.
\looseness=-1\relax\par
\begin{equation}
\begin{aligned}
 j_i^*&=\argmin_{j\in\mathcal N_i}d(\hat\tau_{ij},h_j),\\
 \ell_i&=\frac{10}{3H}\|\hat v_{ij_i^*,t}-v_{ij_i^*,t}\|_F^2
 +20d(\hat\tau_{ij_i^*},h_{j_i^*}).
\end{aligned}
\label{eq:wta}
\end{equation}
Every matched proposal receives denoising and geometric supervision for its selected alternative. Unmatched proposals use themselves as targets, yielding zero residuals. The loss $\mathcal L_D$ sums the matched-proposal mean and 0.05 times the unmatched-proposal mean, with empty groups contributing zero. The weaker term discourages unnecessary corrections.
\looseness=-1\relax\par

\noindent\textbf{Shared JEPA conditioning.} Refinement and scoring require temporal traffic context without online VLM inference. We therefore adapt a JEPA encoder $E_\phi$ initialized from Drive-JEPA~\cite{wang2026drivejepa} through masked video-feature prediction. Given visible clip context, a predictor regresses layer-normalized features of an exponential moving average (EMA) encoder at masked positions in eight-frame clips sampled at 2\,Hz. The objective $\mathcal L_J$ averages L1 error across feature dimensions, masked positions, and two mask families. Only this objective updates the encoder and predictor. For planning, a separate pass encodes the previous and current front-camera frames $I_s^{\rm obs}$. Independent adapters $A_D$ and $A_S$ combine the shared features with ego history $h_s^{\rm ego}$.
\looseness=-1\relax\par
\begin{equation}
\begin{aligned}
 F_s&=\sg[E_\phi(I_s^{\rm obs})],\\
 C_s^D&=A_D(F_s,h_s^{\rm ego}),\qquad
 C_s^S=A_S(F_s,h_s^{\rm ego}),
\end{aligned}
\label{eq:video_conditions}
\end{equation}
Here $\sg$ blocks planning gradients. Independent adapters fuse shared video features with ego history for DiT and gated scorer cross-attention. Both modules reuse one observed-frame encoding. Adaptation clips may include future frames, whereas planning uses observed frames during training and inference. Adapter, masking, and EMA details are given in the supplement.
\looseness=-1\relax\par

\ifdefined\arxivtextspacing\enlargethispage{\baselineskip}\fi
\subsection{Training objectives and inference}
\label{sec:optimization}
Refinement can change execution outcomes. We generate the refined pool $Q_s$ from all 64 proposals using target-free, 12-step DDIM sampling~\cite{song2021ddim} and recompute evaluator labels for $\mathcal L_S(Q_s)$. These samples train scoring, while target-derived noisy residuals supervise DiT. Calibration supervises individual quality estimates, while candidate selection depends on relative ordering. We therefore add an optional pairwise ranking objective. We retain native pairs with evaluator score gap $w_{ij}=J_i-J_j>0.005$ and at least one candidate among the top eight under predicted or evaluator scores.
\looseness=-1\relax\par
\begingroup\small\thinmuskip=1mu\medmuskip=2mu\thickmuskip=2mu\def\mintagsep{2pt}
\begin{equation}
 \mathcal L_R(C_s)=\frac{\sum_{(i,j)}w_{ij}\log\left(1+
 \exp[-(\Jhat_i-\Jhat_j)/T]\right)}{\sum_{(i,j)}w_{ij}}.
\label{eq:ranking}
\end{equation}
\endgroup
Here $T=0.05$ and empty pair sets contribute zero. The loss weights pairs by quality gap and encourages score separation even when their ordering is correct. For PDMS training, both stages use $\lambda_R=0$ on real data and $0.05$ on mixed data (\Cref{sec:ablations}).
\looseness=-1\relax\par

The objectives combine generation and scoring, adding residual refinement and video adaptation in Stage~2.
\begin{equation}
\begin{aligned}
 \mathcal L^{(1)}&=\mathcal L_G+\mathcal L_S(P_s)+\lambda_Rr(e)\mathcal L_R(P_s),\\
 \mathcal L^{(2)}&=\mathcal L_G+\mathcal L_D+\mathcal L_S(Q_s)\\
 &\quad+\mathcal L_J+\lambda_Rr(e)\mathcal L_R(Q_s).
\end{aligned}
\label{eq:total}
\end{equation}
Planning losses average scenes equally, and video regression averages valid clips globally. Refiner and scorer gradients update shared scene features with parent and scored coordinates detached. Planning adapters receive detached video features. Separate gradient clipping prevents planning gradients from rescaling video updates.
\looseness=-1\relax\par

At inference, we select the trajectory maximizing the predicted aggregate score over $P_s$ in Stage~1 or $Q_s$ in Stage~2. Stage~2 shares one observed-frame JEPA encoding across refinement and scoring. The archive, evaluator labels, future frames, JEPA predictor, and EMA encoder are required only for training.
\looseness=-1\relax\par

%% file: sections/04_experiments.tex
\section{Experiments}
\label{sec:experiments}

\subsection{Experimental setup}
We first evaluate selected plans against logged traffic in NAVSIM v1 and then test repeated control in HUGSIM. Candidate diagnostics separate generation quality from selection quality. NAVSIM v2 further evaluates consistency across successive plans and performance when other road users respond.
\looseness=-1\relax\par

\noindent\textbf{Data and models.} \method{} trains on 103,263 NAVSIM Navtrainval scenes after excluding 25 unqualified trajectory pools~\cite{dauner2024navsim}. \method{}-Scale adds 129,409 SimScale scenes, for a total of 232,672~\cite{simscale2025}. The counts refer to planner training and exclude external pretraining. Stage~1 (S1) selects from 64 autoregressive proposals using their predicted scores. Stage~2 (S2) refines all 64 proposals with 12 deterministic DDIM steps and scores the refined trajectories before selection.
\looseness=-1\relax\par

\noindent\textbf{Training and reporting.} We train on eight H20 GPUs with AdamW, batch size 48, and one training seed per configuration. For v1 PDMS and HUGSIM, \method{} omits the pairwise ranking loss and Scale includes it. Both use ranking in the NAVSIM v2 evaluations. We report the best complete evaluation on the reported benchmark per run. For v2, checkpoints are selected separately on Navtest and Navhard. HUGSIM uses the Navtest PDMS-selected checkpoints and a common controller for all four configurations. We quote published benchmark scores for other planners and measure latency locally. The supplement gives training schedules and the number of checkpoint evaluations per run.
\looseness=-1\relax\par

\subsection{Planning quality and inference cost}
\input{sections/04_primary_tables}
\noindent\textbf{NAVSIM v1.} On 12,146 Navtest scenes, \method{} achieves \FinalBest{} PDMS, the highest score among the compared planners trained only on NAVSIM planning data (\cref{tab:main}). Relative to ChainFlow-VLA, \method{} S2 makes more progress and improves drivable-area compliance. The two planners have the same no-at-fault collision score, while ChainFlow-VLA has a higher time-to-collision score. With expanded data, \method{}-Scale S2 reaches \ScaledRefinedBest{} PDMS and has the highest ego progress in the table. Within each data regime, refinement and rescoring improve the selected plan over S1.
\looseness=-1\relax\par

\input{sections/04_hugsim_table}
\noindent\textbf{Closed-loop HUGSIM.} Each executed plan changes the next observation in HUGSIM, testing whether the Navtest gains carry over to repeated planning. Across 345 matched HUGSIM episodes~\cite{zhou2024hugsim}, \method{} S2 leads the standard NAVSIM group in mean route completion (55.2) and HD-Score (44.0). With expanded training, \method{}-Scale S2 achieves the highest averages in the full comparison (56.1 and 46.7, respectively). HD-Score combines the fraction of the route completed with execution quality along that route (\cref{tab:hugsim}). Refinement raises mean HD-Score by 7.17 points on real training data and 2.72 on mixed data, with the largest gains on Medium and Hard episodes, respectively. In the Scale model, average collision avoidance and time-to-collision scores both improve after refinement, while comfort falls from 99.6 to 95.0. On Extreme episodes, refinement reduces both route completion and HD-Score in both training regimes.
\looseness=-1\relax\par

\input{sections/04_latency_table}
\noindent\textbf{Inference cost.} Shared observed-frame features avoid repeated encoding during refinement and scoring. We compare expanded-data checkpoints with ChainFlow-VLA on identical Navtest scenes using one scene per H20 worker, 64 candidates, and 12 steps for both refined planners (\cref{tab:inference_latency}). Both training regimes share the inference architecture at each stage. Mean forward-pass time falls from 1,295 to 349\,ms, a 73\% reduction. Timing covers GPU-resident inputs to the selected plan, excluding preprocessing and simulator evaluation, and compares complete implementations. Within Scale, refinement and shared video conditioning cost 200\,ms over S1 for a 0.46-point PDMS gain.
\looseness=-1\relax\par

\ifdefined\arxivtextspacing
\input{sections/04_proposal_table}
\input{sections/04_refinement_table}
\input{sections/04_scoring_diagnostics}
\input{sections/04_epdms_tables}
\fi
\input{sections/04_score_distribution}
\subsection{Learning from scored alternatives}
\label{sec:ablations}
\noindent\textbf{What supervision does one scene provide?} A high-scoring demonstration gives little indication of how nearby alternatives will execute. In \cref{fig:census}, the reference scores remain high across much of the real archive even as the fraction of successful deformations varies widely across scenes. Overall, 49.17\% of real candidate paths score 100 and 34.05\% score zero. The synthetic archive also spans the score range. Its distribution is computed on a separate census of 204,831 scenes.
\looseness=-1\relax\par

Screening the real archive yields 2.61M geometric targets across 95,998 scenes, about 27 per scene with retained targets. In 55.45\% of these scenes, at least one retained alternative outscores the human reference under the evaluator. Across scenes with retained targets, the best target improves on the reference by 4.65 PDMS on average (95\% log-bootstrap interval $[4.38,4.94]$). For almost all references below 95 PDMS, the weighted score loss comes from progress alone. This leaves room to improve progress while retaining the safety and tracking checks used to screen targets. The 7,265 scenes without screened targets still contribute expert and score supervision.
\looseness=-1\relax\par

\input{sections/04_qualitative_figure}
\ifdefined\arxivtextspacing\else
\input{sections/04_proposal_table}
\fi
\noindent\textbf{Proposal generation and selection.} To see whether the generator supplies better options, we measure Oracle, the best evaluator score among the 64 proposals in each scene. Regret is the difference between Oracle and the score of the selected proposal. All rows of \cref{tab:ablation_stage1} retain expert supervision and evaluator labels for the planner's own proposals. Holding the scoring losses fixed, matching screened targets to distinct proposals (Dense) raises Oracle by 0.49 points and selected PDMS by 0.64 on real data. The Oracle gain shows that a higher-quality plan is available before selection.
\looseness=-1\relax\par

The scoring objectives affect which proposal is chosen. With Dense and Rank fixed, replaying archived trajectories lowers selected PDMS by 0.11 on real data with little change in Oracle. With Dense and Replay fixed, pairwise ranking lowers selected PDMS by 0.34 on real data but raises it by 0.26 on mixed data, again with nearly unchanged Oracle. This difference motivates the ranking-loss settings used for the main v1 results.
\looseness=-1\relax\par

\ifdefined\arxivtextspacing\else
\input{sections/04_refinement_table}
\fi
\noindent\textbf{Local residual supervision.} The configurations in \cref{tab:ablation_stage2} share an S1 initialization trained with Rank. The Screen configuration raises Oracle from 96.94 to 99.77 and selected PDMS from 93.24 to 94.75 over the expert-target control. The PDMS gain has a paired 95\% log-bootstrap interval of $[0.62,2.42]$ at the selected checkpoints. The larger Oracle gain widens regret from 3.70 to 5.03. Using up to four compatible targets with reconstruction-based assignment yields 94.87 selected PDMS and 99.56 Oracle, reducing regret to 4.69.
\looseness=-1\relax\par

\ifdefined\arxivtextspacing\else
\input{sections/04_scoring_diagnostics}
\fi
\ifdefined\arxivtextspacing\else
\input{sections/04_epdms_tables}
\fi
\noindent\textbf{Why rescore after refinement?} Following the same candidates through denoising on 128 scenes, mean candidate PDMS rises from 75.16 to 81.05. Refining the highest-scoring initial proposal yields 89.49 against a refined Oracle of 99.77. The preferred index can therefore change after denoising. On fixed candidate pools (\cref{tab:scorer_diagnostics}), the S1 scorer selects better initial proposals (94.42 vs.\ 93.86 PDMS), while the S2 scorer selects better refined trajectories (94.72 vs.\ 93.53). Each scorer performs better on candidates from its training stage.
\looseness=-1\relax\par

\noindent\textbf{Shared observed-frame context.} Enabling the shared conditioning branch in both refiner and scorer raises selected PDMS from 93.94 to 94.87, with nearly unchanged Oracle. Adapting JEPA further raises it to 94.93 (\cref{tab:ablation_stage2}). Both modules combine shared observed-frame features with ego history through independent adapters. Restoring observed video to the S2 scorer on fixed candidate pools improves selected PDMS by 4.41 points on initial proposals and 3.20 on refined trajectories (\cref{tab:scorer_diagnostics}).
\looseness=-1\relax\par

\noindent\textbf{Driving examples.} The examples come from a pool of 41 Navtest scenes in which \method{}-Scale S2 scores at least 95 PDMS and each compared baseline scores below 80. In the crossing-vehicle case (\cref{fig:qualitative_collision}, bottom) and the crossing-pedestrian case (\cref{fig:qualitative_collision}, top), Scale S2 preserves clearance while FPD-Drive-Pro and ChainFlow-VLA collide under the evaluator. In \cref{fig:teaser}, Scale S2 takes a lateral path around the vehicles ahead. FPD-Drive-Pro collides, and ChainFlow-VLA avoids contact but fails the time-to-collision criterion. Warmer regions in the maps indicate higher refiner attention to road users and nearby road within each image.
\looseness=-1\relax\par

\ifdefined\arxivmainflow\enlargethispage{\baselineskip}\fi
\subsection{Temporal consistency and reactive traffic}
Both v2 evaluations use S1 models trained with the original six-component PDMS objective and ranking. We recompute EPDMS after human-reference filtering~\cite{cao2025pseudo}. Extended comfort measures consistency between successive plans beyond individual trajectory execution.
\looseness=-1\relax\par

\noindent\textbf{Navtest.} On the same 12,146 scenes, \method{} and Scale score 91.15 and 91.32 EPDMS, below WA-JEPA's 91.7 (\cref{tab:navtest_epdms}). Expanded training improves progress while reducing history and extended comfort. Higher individual-plan scores therefore need not imply consistency between successive plans.
\looseness=-1\relax\par

\noindent\textbf{Navhard.} On 450 original scenes and 5,462 rendered reactive continuations, Scale raises EPDMS from 50.80 to 56.13, second among compared planners behind DriveZero-Scale (57.1). Continuation no-at-fault collision and time-to-collision scores improve by 4.7 and 4.1 points over real-only training, with little change on original observations (\cref{tab:navhard}). Extended comfort falls from 70.6 to 32.3 on continuations. Across both v2 evaluations, expanded training reduces cross-plan comfort, which individual trajectory supervision does not constrain.
\looseness=-1\relax\par

%% file: sections/04_primary_tables.tex
\begin{table}[!t]
\centering
\captionsetup{skip=3pt,justification=justified,singlelinecheck=false}
\caption{\textbf{Navtest PDMS.} Published baselines. Data counts planning scenes, excluding pretraining. Bold/underline mark best/second PDMS per data regime, with displayed ties.}
\label{tab:main}
\fontsize{8.0}{9.2}\selectfont
\setlength{\tabcolsep}{2.3pt}\renewcommand{\arraystretch}{0.936625}
\resizebox{\linewidth}{!}{%
\begin{tabular}{lcrrrrrr}
\toprule
\rowcolor{TableHeader}
Method & Data & NC & DAC & TTC & Comf. & EP & PDMS $\uparrow$\\
\midrule
\rowcolor{TableGroup}
\multicolumn{8}{l}{\textit{End-to-end planning ($\sim$103k scenes)}}\\
TransFuser~\cite{dauner2024navsim} & 103k & 97.7 & 92.8 & 92.8 & 100.0 & 79.2 & 84.0\\
\rowcolor{TableAlt}
DiffusionDrive~\cite{diffusiondrive2024} & 103k & 98.2 & 96.2 & 94.7 & 100.0 & 82.2 & 88.1\\
iPad~\cite{ipad2025} & 103k & 98.6 & 98.3 & 94.9 & 100.0 & 88.0 & 91.7\\
\rowcolor{TableAlt}
DriveSuprim~\cite{drivesuprim2025} & 103k & 98.6 & 98.6 & 95.5 & 100.0 & 91.3 & 93.5\\
DrivoR~\cite{kirby2026registers} & 103k & 99.0 & 98.9 & 96.7 & 100.0 & 90.0 & 93.7\\
\rowcolor{TableAlt}
CLOVER~\cite{clover2026} & 103k & 99.1 & 99.0 & 96.9 & 100.0 & 91.7 & 94.5\\
DriveZero~\cite{drivezero2026} & 100k & 99.0 & 99.2 & 96.0 & 100.0 & 93.1 & \underline{94.8}\\
\midrule
\rowcolor{TableGroup}
\multicolumn{8}{l}{\textit{World models and video representations ($\sim$103k scenes)}}\\
World4Drive~\cite{world4drive2025} & 103k & 97.4 & 94.3 & 92.8 & 100.0 & 79.9 & 85.1\\
\rowcolor{TableAlt}
DriveFuture~\cite{drivefuture2026} & 103k & 98.8 & 99.1 & 95.4 & 100.0 & 84.2 & 90.7\\
Auto-JEPA~\cite{autojepa2026} & 103k & 98.4 & 98.3 & 95.0 & 100.0 & 87.1 & 91.3\\
\rowcolor{TableAlt}
WA-JEPA~\cite{wajepa2026} & 103k & 99.5 & 98.3 & 97.7 & 100.0 & 85.0 & 91.8\\
Drive-JEPA~\cite{wang2026drivejepa} & 103k & 99.1 & 98.2 & 96.2 & 99.9 & 91.4 & 93.7\\
\midrule
\rowcolor{TableGroup}
\multicolumn{8}{l}{\textit{Vision-language-action models ($\sim$103k scenes)}}\\
UniVLA~\cite{univla2025} & 103k & 96.9 & 91.1 & 91.7 & 96.7 & 76.8 & 81.7\\
\rowcolor{TableAlt}
DriveVLA-W0~\cite{drivevlaw02025} & 103k & 98.7 & 99.1 & 95.3 & 99.3 & 83.3 & 90.2\\
ReCogDrive~\cite{recogdrive2025} & 103k & 97.9 & 97.3 & 94.9 & 100.0 & 87.3 & 90.8\\
\rowcolor{TableAlt}
FLARE-4B~\cite{flare2026} & 103k & 98.5 & 98.4 & 96.0 & 100.0 & 86.0 & 91.4\\
ChainFlow-VLA~\cite{wang2026chainflow} & 103k & 99.2 & 99.0 & 97.2 & 99.9 & 91.9 & \underline{94.8}\\
\midrule
\rowcolor{TableOurs}
\textbf{DriveSTA S1} & 103k & 99.1 & 99.0 & 96.5 & 100.0 & 92.6 & \underline{94.8}\\
\rowcolor{TableOurs}
\textbf{DriveSTA S2} & 103k & 99.2 & 99.2 & 96.8 & 100.0 & 92.9 & \textbf{95.1}\\
\midrule
\rowcolor{TableGroup}
\multicolumn{8}{l}{\textit{Expanded training data ($>$103k scenes)}}\\
AutoVLA~\cite{autovla2025} & 185.3k & 98.4 & 95.6 & 98.0 & 99.9 & 81.9 & 89.1\\
\rowcolor{TableAlt}
CLEAR~\cite{clear2026} & 130k & 99.1 & 98.8 & 97.2 & 99.6 & 89.7 & 93.7\\
RAP-DINO~\cite{rap2025} & 565.5k & 99.1 & 98.9 & 96.7 & 100.0 & 90.3 & 93.8\\
\rowcolor{TableAlt}
DrivoR-Scale~\cite{kirby2026registers} & 237k & 99.1 & 99.2 & 96.9 & 100.0 & 91.6 & 94.6\\
FPD-Drive~\cite{chen2026fpd} & 240k & 99.2 & 99.5 & 97.5 & 99.9 & 91.3 & 94.9\\
\rowcolor{TableAlt}
DriveZero-Scale~\cite{drivezero2026} & 337k & 99.2 & 99.4 & 97.3 & 99.9 & 92.5 & 95.3\\
FPD-Drive-Pro~\cite{chen2026fpd} & 398.8k & 99.3 & 99.3 & 97.4 & 100.0 & 93.1 & \underline{95.5}\\
\midrule
\rowcolor{TableOurs}
\textbf{DriveSTA-Scale S1} & 232.7k & 99.0 & 99.2 & 96.6 & 100.0 & 93.0 & 95.1\\
\rowcolor{TableOurs}
\textbf{DriveSTA-Scale S2} & 232.7k & 99.3 & 99.1 & 97.4 & 100.0 & 93.4 & \textbf{95.6}\\
\bottomrule
\end{tabular}%
}
\end{table}

%% file: sections/04_hugsim_table.tex
\begin{table}[!t]
\centering
\captionsetup{skip=3pt,justification=justified,singlelinecheck=false}
\caption{\textbf{HUGSIM closed-loop results.} E/M/H/X denote Easy/Medium/Hard/Extreme. Bold/underline mark best/second averages within each training-data group.}
\label{tab:hugsim}
\fontsize{7}{8.5}\selectfont
\setlength{\tabcolsep}{1.2pt}\renewcommand{\arraystretch}{0.97}
\resizebox{\linewidth}{!}{%
\begin{tabular}{@{}lc|rrrrr|rrrrr@{}}
\toprule
\rowcolor{TableHeader}
 & & \multicolumn{5}{c|}{RC $\uparrow$} & \multicolumn{5}{c}{HD-Score $\uparrow$}\\
\rowcolor{TableHeader}
Method & Data & E & M & H & X & Avg. & E & M & H & X & Avg.\\
\midrule
\rowcolor{TableGroup}
\multicolumn{12}{l}{\textit{Standard NAVSIM training ($\sim$103k scenes or a subset)}}\\
DrivoR-Reg$^\dagger$~\cite{rowe2026gigapixel} & 103k & 75.8 & 28.0 & 32.3 & 46.8 & 40.5 & 60.3 & 8.0 & 11.5 & 25.8 & 20.7\\
\rowcolor{TableAlt}
LTF$^\dagger$~\cite{chitta2023transfuser} & 103k & 67.8 & 35.1 & 26.2 & 40.5 & 38.9 & 58.9 & 18.0 & 9.8 & 25.9 & 23.7\\
DrivoR~\cite{kirby2026registers} & 103k & 80.9 & 50.5 & 33.8 & 47.1 & 49.8 & 73.3 & 34.6 & 18.8 & 32.5 & 35.7\\
\rowcolor{TableAlt}
SimVTA$^\dagger$~\cite{jing2026simvta} & 103k & 78.8 & 57.6 & 38.3 & 40.1 & 51.5 & 69.9 & 45.0 & 23.3 & 27.2 & 38.8\\
DriveZero~\cite{drivezero2026} & 100k & 85.5 & 63.7 & 31.7 & 43.3 & \underline{53.9} & 78.5 & 47.2 & 16.0 & 28.7 & \underline{39.4}\\
\midrule
\rowcolor{TableOurs}
\textbf{\method{} S1} & 103k & 83.1 & 56.9 & 37.1 & 41.2 & 51.9 & 73.8 & 41.0 & 20.7 & 25.7 & 36.9\\
\rowcolor{TableOurs}
\textbf{\method{} S2} & 103k & 84.3 & 68.0 & 35.2 & 39.5 & \textbf{55.2} & 75.9 & 58.6 & 23.0 & 25.3 & \textbf{44.0}\\
\midrule
\rowcolor{TableGroup}
\multicolumn{12}{l}{\textit{Expanded training data ($>$103k scenes)}}\\
Gigapixel-Reg~\cite{rowe2026gigapixel} & 335k & 84.1 & 54.3 & 29.8 & 41.1 & 49.2 & 65.8 & 38.8 & 13.1 & 24.6 & 33.2\\
\rowcolor{TableAlt}
DrivoR-Scale~\cite{rowe2026gigapixel} & 237k & 70.4 & 54.4 & 28.2 & 39.2 & 46.4 & 64.8 & 50.3 & 16.8 & 25.8 & 38.1\\
Gigapixel~\cite{rowe2026gigapixel} & 335k & 78.6 & 60.7 & 32.7 & 35.4 & 50.1 & 67.4 & 51.9 & 19.1 & 21.6 & 38.5\\
\rowcolor{TableAlt}
FPD-Drive~\cite{chen2026fpd} & 240k & 86.2 & 62.8 & 38.2 & 43.3 & \underline{55.3} & 78.5 & 57.0 & 25.2 & 28.1 & 45.1\\
DriveZero-Scale~\cite{drivezero2026} & 337k & 87.5 & 64.9 & 34.6 & 39.8 & 54.5 & 81.9 & 60.5 & 24.7 & 27.6 & \underline{46.6}\\
\midrule
\rowcolor{TableOurs}
\textbf{\method{}-Scale S1} & 232.7k & 83.2 & 63.1 & 33.4 & 44.6 & 54.0 & 74.0 & 56.8 & 21.4 & 30.3 & 44.0\\
\rowcolor{TableOurs}
\textbf{\method{}-Scale S2} & 232.7k & 84.5 & 67.9 & 37.8 & 40.6 & \textbf{56.1} & 75.2 & 61.5 & 27.6 & 27.3 & \textbf{46.7}\\
\bottomrule
\end{tabular}%
}
\end{table}

%% file: sections/04_latency_table.tex
\begin{table}[!t]
\centering
\captionsetup{skip=3pt,justification=justified,singlelinecheck=false}
\caption{\textbf{PDMS and inference latency.} Matched evaluation on 12,146 Navtest scenes with H20 and batch~1. DriveSTA timings use 232.7k checkpoints.}
\label{tab:inference_latency}
\fontsize{7.5}{9}\selectfont
\setlength{\tabcolsep}{1pt}\renewcommand{\arraystretch}{1.12}
\begin{tabular*}{\linewidth}{@{\extracolsep{\fill}}lrrrrrr@{}}
\toprule
\rowcolor{TableHeader}
 & \multicolumn{2}{c}{PDMS $\uparrow$} & \multicolumn{3}{c}{Latency (ms) $\downarrow$} & \\
\cmidrule(lr){2-3}\cmidrule(lr){4-6}
\rowcolor{TableHeader}
Architecture & $\sim$103k & Expanded & Mean $\pm$ SD & Median & P95 & Hz $\uparrow$\\
\midrule
ChainFlow-VLA & 94.85 & --- & $1294.68\pm4.47$ & 1293.81 & 1302.79 & 0.77\\
\rowcolor{TableOurs}
\method{} S1 & 94.76 & 95.09 & $148.99\pm0.85$ & 148.91 & 150.32 & 6.71\\
\rowcolor{TableOurs}
\method{} S2 & 95.14 & 95.55 & $349.30\pm5.21$ & 347.89 & 359.29 & 2.86\\
\bottomrule
\end{tabular*}
\end{table}

%% file: sections/04_proposal_table.tex
\begin{table}[!htb]
\centering
\captionsetup{skip=3pt,justification=justified,singlelinecheck=false}
\caption{\textbf{Stage~1 supervision.} Dense matches screened targets, Replay samples archives, and Rank supervises ordering. Zero is the zero-score percentage.}
\label{tab:ablation_stage1}
\fontsize{8.0}{9.2}\selectfont
\setlength{\tabcolsep}{0pt}\renewcommand{\arraystretch}{1.08}
\begin{tabular}{@{}*{3}{>{\centering\arraybackslash}p{0.12\columnwidth}}*{4}{>{\centering\arraybackslash}p{0.16\columnwidth}}@{}}
\toprule
\rowcolor{TableHeader}
Dense & Replay & Rank & PDMS$\uparrow$ & Oracle$\uparrow$ & Regret$\downarrow$ & Zero$\downarrow$\\
\midrule
\rowcolor{TableGroup}
\multicolumn{7}{l}{\textit{DriveSTA: 103.3k real scenes}}\\
$\times$ & $\times$ & $\times$ & 93.92 & 99.31 & 5.39 & 2.19\\
$\times$ & $\checkmark$ & $\checkmark$ & 93.78 & 99.29 & 5.51 & 2.16\\
$\checkmark$ & $\times$ & $\checkmark$ & 94.53 & 99.77 & 5.23 & 2.03\\
$\checkmark$ & $\checkmark$ & $\checkmark$ & 94.42 & 99.78 & 5.36 & 2.15\\
\midrule
\rowcolor{TableOurs}
$\checkmark$ & $\checkmark$ & $\times$ & 94.76 & 99.75 & 4.99 & 1.86\\
\midrule
\rowcolor{TableGroup}
\multicolumn{7}{l}{\textit{Scale: 103.3k real + 129.4k synthetic scenes}}\\
$\checkmark$ & $\checkmark$ & $\times$ & 94.83 & 99.77 & 4.94 & 1.81\\
\rowcolor{TableOurs}
$\checkmark$ & $\checkmark$ & $\checkmark$ & 95.09 & 99.78 & 4.68 & 1.65\\
\bottomrule
\end{tabular}
\end{table}

%% file: sections/04_refinement_table.tex
\begin{table}[!htb]
\centering
\captionsetup{skip=3pt,justification=justified,singlelinecheck=false}
\caption{\textbf{Stage~2 configurations (real data, +Rank).} Screen combines screened targets, proximity gates, and a zero-residual fallback. $J_D/J_S$ enable refiner/scorer conditioning, including video, ego history, and fusion layers. Adapt updates JEPA. $K$ caps compatible targets; assignment precedes noise for $K=1$ and follows reconstruction for $K=4$.}
\label{tab:ablation_stage2}
\fontsize{8.0}{9.2}\selectfont
\setlength{\tabcolsep}{0pt}\renewcommand{\arraystretch}{1.08}
\begin{tabular}{@{}>{\centering\arraybackslash}p{0.095\columnwidth}*{2}{>{\centering\arraybackslash}p{0.065\columnwidth}}>{\centering\arraybackslash}p{0.095\columnwidth}>{\centering\arraybackslash}p{0.05\columnwidth}*{4}{>{\centering\arraybackslash}p{0.1575\columnwidth}}@{}}
\toprule
\rowcolor{TableHeader}
Screen & $J_D$ & $J_S$ & Adapt & $K$ & PDMS$\uparrow$ & Oracle$\uparrow$ & Regret$\downarrow$ & Zero$\downarrow$\\
\midrule
$\times$ & $\checkmark$ & $\checkmark$ & $\times$ & 1 & 93.24 & 96.94 & 3.70 & 1.84\\
$\checkmark$ & $\checkmark$ & $\checkmark$ & $\times$ & 1 & 94.75 & 99.77 & 5.03 & 1.89\\
$\checkmark$ & $\times$ & $\times$ & $\times$ & 4 & 93.94 & 99.55 & 5.61 & 2.92\\
$\checkmark$ & $\checkmark$ & $\checkmark$ & $\times$ & 4 & 94.87 & 99.56 & 4.69 & 2.31\\
\midrule
\rowcolor{TableOurs}
$\checkmark$ & $\checkmark$ & $\checkmark$ & $\checkmark$ & 4 & 94.93 & 99.61 & 4.68 & 2.27\\
\bottomrule
\end{tabular}
\end{table}

%% file: sections/04_scoring_diagnostics.tex
\begin{table}[!htb]
\centering
\captionsetup{skip=3pt,justification=justified,singlelinecheck=false}
\caption{\textbf{Selection on fixed candidate pools.} Upper rows compare S1/S2 selected PDMS. Lower rows give gains from restoring S2 scorer video with weights, candidates, and nonvisual inputs fixed; refiner video is unchanged. Paired 95\% intervals resample 44 driving logs at the selected checkpoints.}
\label{tab:scorer_diagnostics}
\small
\setlength{\tabcolsep}{4pt}
\begin{tabular*}{\linewidth}{@{\extracolsep{\fill}}lrr@{}}
\toprule
\rowcolor{TableHeader}
& Initial proposals & Refined trajectories\\
\midrule
Stage~1 scorer & \textbf{94.42} & 93.53\\
Stage~2 scorer & 93.86 & \textbf{94.72}\\
\midrule
Observed $-$ masked & $+4.41$ & $+3.20$\\
95\% interval & $[3.81,4.98]$ & $[2.57,3.78]$\\
\bottomrule
\end{tabular*}
\end{table}

%% file: sections/04_epdms_tables.tex
\begin{table*}[!t]
\centering
\captionsetup{skip=3pt,justification=justified,singlelinecheck=false}
\caption{\textbf{Navhard EPDMS.} DriveSTA S1 uses corrected human-reference filtering. Bold/underline mark best/second-best EPDMS.}
\label{tab:navhard}
\fontsize{7.5}{8.6}\selectfont
\setlength{\tabcolsep}{0pt}\renewcommand{\arraystretch}{1.0287}
\begin{tabular}{@{}>{\raggedright\arraybackslash}p{0.155\textwidth}>{\centering\arraybackslash}p{0.059\textwidth}*{18}{>{\centering\arraybackslash}p{0.04\textwidth}}>{\centering\arraybackslash}p{0.066\textwidth}@{}}
\toprule
\rowcolor{TableHeader}
 & & \multicolumn{9}{c}{Original observations} & \multicolumn{9}{c}{Rendered continuations} & \\
\cmidrule(lr){3-11}\cmidrule(lr){12-20}
\rowcolor{TableHeader}
\multirow{-2}{*}{Method} & \multirow{-2}{*}{Data} & NC & DAC & DDC & TL & EP & TTC & LK & HC & EC & NC & DAC & DDC & TL & EP & TTC & LK & HC & EC & \multirow{-2}{*}{EPDMS $\uparrow$}\\

\midrule
IDOL~\cite{zhang2026idol} & 103k & 97.2 & 89.6 & 98.0 & 99.6 & 82.3 & 96.9 & 95.3 & 97.6 & 72.9 & 84.9 & 81.7 & 87.7 & 98.5 & 84.7 & 81.1 & 50.0 & 95.4 & 64.0 & 38.0\\
\rowcolor{TableAlt}
SpanVLA~\cite{zhou2026spanvla} & 100k & 99.1 & 94.2 & 99.8 & 100.0 & 66.6 & 98.4 & 91.8 & 97.3 & 47.1 & 89.8 & 84.5 & 94.8 & 98.6 & 63.4 & 88.2 & 53.1 & 95.2 & 50.8 & 40.8\\
GuideFlow~\cite{liu2025guideflow} & 103k & 98.8 & 95.5 & 99.1 & 99.5 & 76.0 & 99.1 & 94.4 & 97.5 & 52.4 & 88.8 & 89.4 & 89.4 & 98.8 & 80.3 & 86.7 & 52.9 & 96.9 & 56.9 & 43.0\\
\rowcolor{TableAlt}
ZTRS~\cite{ztrs2025} & 103k & 99.1 & 98.9 & 100.0 & 100.0 & 65.6 & 98.9 & 94.4 & 95.8 & 38.2 & 91.2 & 94.4 & 96.3 & 98.7 & 59.4 & 90.5 & 57.3 & 97.5 & 63.0 & 46.2\\
CLOVER~\cite{clover2026} & 103k & 99.4 & 98.0 & 99.9 & 100.0 & 75.7 & 98.9 & 94.4 & 97.1 & 56.0 & 89.8 & 88.2 & 93.2 & 99.0 & 74.2 & 87.7 & 52.6 & 97.0 & 57.3 & 48.3\\
\rowcolor{TableAlt}
DrivoR~\cite{kirby2026registers} & 85k & 98.8 & 95.1 & 98.9 & 100.0 & 72.6 & 98.7 & 94.0 & 97.6 & 73.3 & 90.2 & 88.4 & 91.9 & 98.6 & 70.0 & 88.0 & 50.1 & 98.5 & 76.2 & 48.3\\
Gigapixel~\cite{rowe2026gigapixel} & 335k & 99.4 & 95.8 & 99.4 & 99.8 & 68.1 & 99.6 & 91.8 & 97.6 & 49.8 & 93.7 & 92.9 & 96.0 & 98.9 & 62.6 & 90.7 & 60.2 & 98.2 & 58.8 & 50.1\\
\rowcolor{TableAlt}
DriveZero~\cite{drivezero2026} & 100k & 98.7 & 98.9 & 99.9 & 99.8 & 79.5 & 98.7 & 95.8 & 97.6 & 51.6 & 90.5 & 89.2 & 94.6 & 98.6 & 81.2 & 88.3 & 57.2 & 94.5 & 41.3 & 51.5\\
DrivoR-Scale~\cite{kirby2026registers} & 219k & 99.1 & 98.2 & 99.3 & 99.8 & 75.4 & 98.7 & 94.9 & 97.6 & 70.2 & 92.3 & 91.6 & 97.3 & 99.1 & 75.7 & 90.6 & 56.1 & 98.4 & 44.7 & 54.6\\
\rowcolor{TableAlt}
DriveFuture~\cite{drivefuture2026} & 103k & 99.8 & 99.8 & 100.0 & 99.6 & 85.7 & 99.8 & 98.7 & 97.6 & 66.2 & 90.6 & 87.5 & 94.1 & 99.1 & 84.6 & 88.8 & 58.3 & 93.5 & 45.6 & 55.5\\
FPD-Drive~\cite{chen2026fpd} & 240k & 99.3 & 98.4 & 99.7 & 99.6 & 78.0 & 99.1 & 96.0 & 97.3 & 50.7 & 94.2 & 93.5 & 95.9 & 99.3 & 79.8 & 92.1 & 59.8 & 94.8 & 41.4 & 55.9\\
\rowcolor{TableAlt}
DriveZero-Scale~\cite{drivezero2026} & 337k & 98.7 & 99.3 & 100.0 & 99.3 & 76.5 & 98.4 & 95.3 & 97.1 & 44.4 & 94.6 & 95.6 & 97.7 & 98.4 & 76.2 & 92.7 & 59.9 & 96.1 & 44.0 & \textbf{57.1}\\
\midrule
\rowcolor{TableOurs}
\textbf{DriveSTA S1} & 103k & 99.2 & 97.3 & 99.9 & 99.8 & 76.4 & 99.3 & 94.7 & 97.1 & 70.2 & 89.7 & 87.3 & 93.6 & 98.9 & 71.6 & 88.3 & 54.6 & 97.7 & 70.6 & 50.8\\
\rowcolor{TableOurs}
\textbf{DriveSTA-Scale S1} & 232.7k & 99.1 & 98.2 & 99.9 & 99.3 & 79.1 & 99.1 & 96.4 & 96.4 & 67.1 & 94.4 & 92.0 & 96.9 & 98.9 & 77.3 & 92.4 & 61.1 & 95.3 & 32.3 & \underline{56.1}\\
\bottomrule
\end{tabular}
\end{table*}

\begin{table}[!htb]
\centering
\captionsetup{skip=3pt,justification=justified,singlelinecheck=false}
\caption{\textbf{Navtest EPDMS.} Corrected human-reference filtering for DriveSTA S1. Bold/underline mark best/second EPDMS.}
\label{tab:navtest_epdms}
\scriptsize
\setlength{\tabcolsep}{1.2pt}\renewcommand{\arraystretch}{1.00}
\resizebox{\linewidth}{!}{%
\begin{tabular}{l*{10}{r}}
\toprule
\rowcolor{TableHeader}
Method & NC & DAC & DDC & TL & EP & TTC & LK & HC & EC & EPDMS $\uparrow$\\
\midrule
EponaV2~\cite{eponav22026} & 98.5 & 97.4 & 99.5 & 99.9 & 87.9 & 98.1 & 97.7 & 98.2 & 77.4 & 88.9\\
\rowcolor{TableAlt}
Auto-JEPA~\cite{autojepa2026} & 98.5 & 98.7 & 98.3 & 99.7 & 90.5 & 97.9 & 94.7 & 97.8 & 75.2 & 89.1\\
DriveFine~\cite{drivefine2026} & 98.7 & 97.3 & 99.5 & 99.8 & 88.7 & 97.8 & 97.7 & 98.4 & 83.8 & 89.7\\
\rowcolor{TableAlt}
DriveFuture~\cite{drivefuture2026} & 98.8 & 99.1 & 99.6 & 99.9 & 86.6 & 98.4 & 96.4 & 98.3 & 74.8 & 89.9\\
SparseDriveV2~\cite{sparsedrivev22026} & 98.1 & 98.1 & 99.6 & 99.8 & 91.1 & 97.3 & 96.9 & 98.2 & 78.4 & 90.1\\
\rowcolor{TableAlt}
CLOVER+anchor~\cite{clover2026} & 99.4 & 99.3 & 99.5 & 99.8 & 86.9 & 98.9 & 95.2 & 98.2 & 75.5 & 90.4\\
SUV~\cite{suv2026} & 99.1 & 97.8 & 99.7 & 99.8 & 87.8 & 98.7 & 98.1 & 98.4 & 88.3 & 91.0\\
\rowcolor{TableAlt}
WA-JEPA~\cite{wajepa2026} & 99.4 & 98.2 & 99.7 & 99.9 & 87.8 & 98.9 & 98.3 & 98.3 & 88.1 & \textbf{91.7}\\
\midrule
\rowcolor{TableOurs}
\textbf{DriveSTA S1} & 99.6 & 99.1 & 99.8 & 99.7 & 89.0 & 99.1 & 95.6 & 97.1 & 75.0 & 91.2\\
\rowcolor{TableOurs}
\textbf{DriveSTA-Scale S1} & 99.5 & 99.2 & 99.8 & 99.6 & 90.1 & 99.2 & 95.7 & 96.0 & 73.2 & \underline{91.3}\\
\bottomrule
\end{tabular}%
}
\end{table}

%% file: sections/04_score_distribution.tex
\begin{figure}[!t]
\centering
\captionsetup{skip=3pt,justification=justified,singlelinecheck=false}
\includegraphics[width=\linewidth]{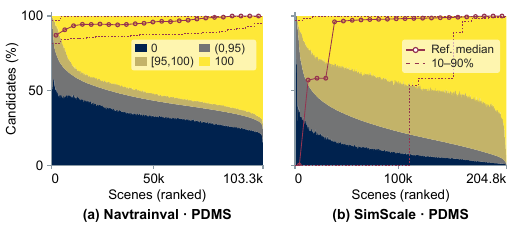}
\caption{\textbf{Trajectory score distributions.} Scenes are ordered by high-score share within each archive. Bands denote score fractions. Reference curves show median PDMS and 10th/90th percentiles.}
\label{fig:census}
\end{figure}

%% file: sections/04_qualitative_figure.tex
\begin{figure}[!t]
\centering
\captionsetup{skip=3pt,justification=justified,singlelinecheck=false}
\includegraphics[width=\linewidth,trim=-1.573661bp 8.5bp 250.328330bp 0,clip]{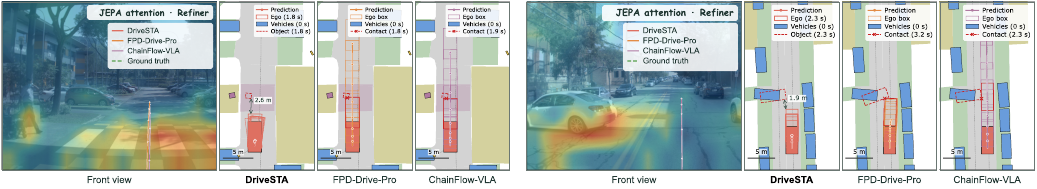}
\par\vspace{4pt}
\includegraphics[width=\linewidth,trim=250.328330bp 0 -1.573661bp 0,clip]{figures/qualitative/figure5_two_cases.pdf}
\caption{\textbf{Navtest examples.} Front-view JEPA attention and bird's-eye trajectories compare \method{}-Scale S2, FPD-Drive-Pro, and ChainFlow-VLA. Red crosses and dashed outlines mark contacts and contacted objects.}
\label{fig:qualitative_collision}
\end{figure}

%% file: sections/05_discussion.tex
\section{Discussion}
\label{sec:discussion}
Improved candidate quality benefits planning only when selection remains accurate. Execution-scored alternatives couple geometric learning with quality estimation, while shared JEPA features support refinement and selection. Search and evaluation remain offline.
\looseness=-1\relax\par

\noindent\textbf{Limitations.} Reference-centered deformations restrict maneuver coverage, while validity remains tied to the evaluator. Logged traffic cannot capture responses to alternative ego plans. Candidate-wise supervision does not directly constrain consistency across consecutive plans, consistent with S1 EPDMS gains alongside reduced temporal comfort. These gaps motivate broader maneuver search, reactive evaluation, and supervision across consecutive plans.
\looseness=-1\relax\par

%% file: sections/06_conclusion.tex
\section{Conclusion}
\label{sec:conclusion}
We present DriveSTA, an end-to-end planner that learns from scored trajectory alternatives within each logged scene. Offline search and execution evaluation supply screened targets for generation and residual refinement and quality labels for selection. Shared observed-frame JEPA features provide temporal context without online VLM inference. DriveSTA leads NAVSIM-only Navtest PDMS, while DriveSTA-Scale leads among compared planners trained on at most 237k scenes. It also achieves the best average HUGSIM RC and HD-Score. Matched Navtest evaluations show lower latency than ChainFlow-VLA. These findings demonstrate the value of recovering supervision beyond recorded maneuvers.
\looseness=-1\relax\par

%% file: sections/08_references.tex
\setlength{\bibsep}{0pt}
\fontsize{9}{10.8}\selectfont

%% file: sections/10_appendix_datasets_metrics.tex
\section{Datasets and evaluation}
\label{app:datasets}
\subsection{Data and observations}
\label{app:exclusions}
NAVSIM~\cite{dauner2024navsim,cao2025pseudo} supplies observations, maps, and evaluator caches. The planner predicts eight ego-frame poses over 4\,s. Four current camera views feed DINO, and the previous/current front views feed JEPA. Four observed ego states provide motion history and navigation. Recorded future traffic supports offline supervision and evaluation.

The real archive contains 103,288 scenes, with 85,109 training and 18,179 validation scenes. We exclude 25 scenes, comprising 20 training and five validation scenes. Ten have initial route or spatial conflicts that future-path changes cannot resolve, and 15 do not yield a complete qualified pool within the search budgets. The retained 103,263 scenes each have 256 high-score candidates. Scenes with no screened targets retain expert and score supervision.

SimScale~\cite{simscale2025} perturbs logged ego states, simulates traffic, renders camera observations, and supplies pseudo-expert references. The reported Scale run retains 129,409 of 134,732 synthetic scenes, giving 232,672 training scenes in total. The larger 204,831-scene archive is used only for the distribution census. Its admission index retains 196,317 scenes, excluding 8,514 for route conflicts (4,215), missing geometry (one), insufficient aggregate score (1,820), failed components (2,175), tracking failures (180), future spatial violations (70), incomplete pools (50), or inconsistent actor tails (three).

Synthetic actor states at 4--5\,s are extrapolated and checked with alternative tails. These states extend the scoring environment without supplying a five-second ego reference. Only the unavailable long-expert loss is masked. Each source uses the evaluator cache associated with its stored labels. Video-window validity is specified in Appendix~\ref{app:video}.

\subsection{NAVSIM evaluation}
\paragraph{PDMS.}
Both Navtest protocols cover 12,146 scenes. NAVSIM v1 evaluates a selected trajectory against logged traffic. For component scores $s_m\in[0,1]$, its aggregate is
\begin{equation}
 J_{\mathrm P}=s_{\mathrm{NC}}s_{\mathrm{DAC}}
 \frac{5s_{\mathrm{EP}}+5s_{\mathrm{TTC}}+2s_{\mathrm C}}{12}.
\label{eq:supp_pdms}
\end{equation}
The product is computed per scene before averaging, and tables report $100J_{\mathrm P}$. Multiplying component means does not recover the aggregate. Driving direction supervises the sixth score head and target admission, with zero weight in the v1 aggregate.

\par\noindent\begin{minipage}{\columnwidth}\captionsetup{hypcap=false}
\centering\suppTableStyle
\captionof{table}{\textbf{Metric components.} ``Product'' denotes a multiplicative penalty. Weights refer to the weighted-average part.}
\label{tab:supp_metrics}
\begin{tabular*}{\columnwidth}{@{\extracolsep{\fill}}llcc@{}}
\toprule\rowcolor{TableHeader}Abbr. & Component & PDMS & EPDMS\\\midrule
NC & No-at-fault collision & Product & Product\\
DAC & Drivable-area compliance & Product & Product\\
DDC & Driving direction & --- & Product\\
TL & Traffic-light compliance & --- & Product\\
EP & Ego progress & 5 & 5\\
TTC & Time-to-collision & 5 & 5\\
C & Comfort & 2 & ---\\
LK & Lane keeping & --- & 2\\
HC & History comfort & --- & 2\\
EC & Extended comfort & --- & 2\\\bottomrule
\end{tabular*}
\end{minipage}\par\medskip

\paragraph{EPDMS.}
Table~\ref{tab:supp_metrics} shows how NAVSIM v2 adds direction and traffic-light penalties and replaces the v1 comfort term with lane keeping, history comfort, and extended comfort. NAVSIM v2 also applies human-reference filtering~\cite{cao2025pseudo}. With filtered scores $\widetilde s_m$ and $\mathcal P=\{\mathrm{NC,DAC,DDC,TL}\}$, it computes
\begin{equation}
 J_{\mathrm E}=\left(\prod_{m\in\mathcal P}\widetilde s_m\right)
 \frac{5\widetilde s_{\mathrm{EP}}+5\widetilde s_{\mathrm{TTC}}+
 2\widetilde s_{\mathrm{LK}}+2\widetilde s_{\mathrm{HC}}+2\widetilde s_{\mathrm{EC}}}{16}.
\label{eq:supp_epdms}
\end{equation}
We use the corrected evaluator that recomputes the aggregate after filtering. EC compares evaluator-paired plans across time. These evaluations retain the six-component PDMS training objective. All four EPDMS runs use Stage~1 with Rank and report the best complete evaluation within each run, separately for Navtest and Navhard. Full-benchmark evaluations have zero errors, and zero-score trajectories remain valid failures.

Navtest uses non-reactive logged traffic and temporal-comfort aggregation, without a synthetic continuation stage. Navhard contains 450 original scenes and 5,462 reactive continuations. For an initial score $J_{1,i}$, continuation score $J_{2,i,l}$, continuation start $x_{i,l}$, and reached endpoint $\hat x_i$, its aggregation is
\begin{equation}
\begin{aligned}
 w_{il}&=\frac{\exp[-\|x_{i,l}-\hat x_i\|^2/(2\sigma^2)]}
 {\sum_j\exp[-\|x_{i,j}-\hat x_i\|^2/(2\sigma^2)]},\\
 J_{\mathrm H,i}&=J_{1,i}\sum_l w_{il}J_{2,i,l},\qquad \sigma^2=0.1\,\mathrm m^2.
\end{aligned}
\label{eq:supp_navhard}
\end{equation}
One checkpoint serves both original and continuation observations within each run. Scores are aggregated per original scene before averaging.

\subsection{HUGSIM evaluation}
HUGSIM~\cite{zhou2024hugsim} updates rendered observations after each executed action. We use the 345-scenario protocol~\cite{rowe2026gigapixel}, comprising 49 Easy, 126 Medium, 86 Hard, and 84 Extreme episodes from KITTI-360 (79), nuScenes (87), PandaSet (51), and Waymo (128). All four \method{} configurations complete the same 345 uniquely matched episodes without evaluation errors. The LTF scores in \cref{tab:hugsim} follow the 345-episode evaluation reported by Rowe et al.~\cite{rowe2026gigapixel}.

DriveSTA/DriveSTA-Scale S1 use the real no-Rank and mixed +Rank checkpoints with 94.7632/95.0932 Navtest PDMS. S2 uses their 95.1424/95.5520 counterparts. Selection coefficients, navigation commands, seed~2, and iLQR settings are shared. Evaluation uses 64 candidates, FP32 with TF32 disabled, replanning every 0.25\,s, and a 50\,ms wall-clock controller budget. The comparison concerns complete training recipes.

For episode $i$ with $T_i$ steps and $\mathrm{RC}_i\in[0,1]$,
\begin{equation}
\begin{aligned}
 P_{it}&=s_{\mathrm{NC},it}s_{\mathrm{DAC},it}
 \frac{5s_{\mathrm{TTC},it}+2s_{\mathrm C,it}}7,\\
 H_i&=\mathrm{RC}_i\,\frac1{T_i}\sum_{t=1}^{T_i}P_{it}.
\end{aligned}
\label{eq:supp_hugsim}
\end{equation}
Tables report $100\overline{\mathrm{RC}}$ and $100\overline H$ over episodes. HUGSIM PDMS averages $100T_i^{-1}\sum_tP_{it}$ across episodes. Overall means weight difficulty levels by their scenario counts, with rounding applied only after aggregation.

\subsection{Visualization protocol}
NAVSIM comparisons reuse saved selected trajectories from DriveSTA-Scale S2 and the public FPD-Drive-Pro, ChainFlow-VLA, and DrivoR-Scale checkpoints, using their native selection rules and identical v1 caches. The selected source pool contains 41 scenes with DriveSTA PDMS$\geq95$ and every baseline below 80. Across the 14 additional cases in Figures~\ref{fig:supp_all_collide} and~\ref{fig:supp_two_collide}, DriveSTA receives NC$=100$ in every case and at least two baselines receive NC$=0$ in 12 cases.

Red crosses and dashed outlines identify evaluator-reported contacts and contacted objects. Clearance annotations measure instantaneous distance at the earliest baseline contact time. They do not guarantee minimum clearance over the full trajectory. These checkpoint-based visualizations are distinct from the published baseline values used in the benchmark tables and do not estimate failure frequency.

Figures~\ref{fig:supp_hugsim_1} and~\ref{fig:supp_hugsim_2} show how DriveSTA-Scale S2 updates its plan as observations change across 16 selected HUGSIM episodes spanning four source datasets and difficulty levels. Each row extracts four equally spaced frames from the same camera view, with an interval of at least 2\,s chosen for the episode duration. Timestamps indicate video time. The images retain the rendered observations and do not constitute a separate performance comparison.

%% file: sections/11_appendix_experimental_details.tex
\section{Implementation and training}
\label{app:details}
\subsection{Trajectory supervision}
\label{app:construction}
\paragraph{Search.}
We construct complete reference-path deformations using Eq.~\ref{eq:warp}, then track and evaluate each path. Initial longitudinal/lateral offset ranges are $[-5,8]$/$[-3,3]$\,m, expanded during recovery to $[-15,40]$/$[-12,12]$\,m. Each 64-query batch allocates 25\% to broad exploration and the remainder to neighborhoods of promising offsets at scales 0.05, 0.15, 0.45, 1.2, and 3\,m. Query budgets increase from 3,072 to 8,192 and then 32,768 as needed. Endpoint keys at 1\,mm resolution avoid repeated queries. The archive retains 3,072 paths across score ranges, independently of query count. A separate admission index collects 256 paths with PDMS at least 95.

\paragraph{Screening.}
Targets require full NC, DAC, TTC, comfort, and driving-direction scores, together with footprint, route, motion, and tracking checks. Limits are 25\,m/s speed, 6\,m/s$^2$ acceleration, 10\,m/s$^3$ jerk, and 4\,m/s$^2$ lateral acceleration. Above 1\,m/s, yaw rate is at most 1\,rad/s and heading--motion discrepancy at most 0.35\,rad. Velocity projected onto the mid-interval heading must be at least $-0.02$\,m/s, integrated backward travel at most 0.01\,m, and initial velocity discrepancy at most 2\,m/s.

Tracking ADE is at most 1\,m, maximum position error at most 2\,m, and heading error at most 0.35\,rad. We visit candidates in descending score order with at most 256 tracking queries, retaining up to 32 targets separated by at least 0.25\,m in whole-path ADE. Padding is masked. Stage~1 rotates up to eight targets per visit. Stage~2 admits up to four nearest targets per parent under the 3\,m ADE, 6\,m endpoint, and 0.5\,rad heading gates. Unmatched parents receive weak zero-residual supervision. These gates apply only to training targets. Inference refines all 64 parents and selects by predicted score without an archive or true-score threshold.

\subsection{Architecture and optimization}
\label{app:configuration}
Tables~\ref{tab:supp_architecture} and~\ref{tab:supp_optim} collect the architecture and optimizer settings. Main Stage~2 models initialize from the same-data Stage~1 checkpoints, retaining no-Rank for DriveSTA and +Rank for Scale. Real-data video controls instead initialize from the +Rank checkpoint. Image LoRA, scene/ego encoding, the autoregressive (AR) generator, diffusion transformer (DiT), scorer, and adapters train jointly. The base DINO weights remain frozen.

\par\noindent\begin{minipage}{\columnwidth}\captionsetup{hypcap=false}
\centering\suppTableStyle
\captionof{table}{\textbf{Model and inference parameters.} Token widths refer to the planning interface.}
\label{tab:supp_architecture}
\begin{tabular*}{\columnwidth}{@{\extracolsep{\fill}}ll@{}}
\toprule
\rowcolor{TableHeader} Parameter & Value \\
\midrule
Current views / resolution & $4$ / $1148\!\times\!672$ \\
Image backbone / LoRA rank & DINOv2 ViT-S/14 / 32 \\
Scene / ego tokens & $64$ / $1$, width 256 \\
AR proposals / output poses & $64$ / $8\!\times\!3$ \\
Horizon / interval & $4$\,s / $0.5$\,s \\
JEPA backbone / width & ViT-L / 1,024 \\
Video tokens: DiT / Scorer & 128 / 512 \\
DiT blocks / heads / width & 12 / 8 / 256 \\
DiT FFN width & 1,024 \\
Prediction / residual scale & $v$ / 1.0 \\
DDIM noise grid / steps & 1,000 / 12 \\
Schedule / sample clipping & Scaled linear / off \\
Metric heads / v1 selection & 6 / PDM argmax \\
Evaluation / noise seed & FP32 / 27 \\
\bottomrule
\end{tabular*}
\end{minipage}\par\medskip

\par\noindent\begin{minipage}{\columnwidth}\captionsetup{hypcap=false}
\centering\suppTableStyle
\captionof{table}{\textbf{Training parameters.} Each Stage~2 planner initializes from Stage~1 trained on the same scene index.}
\label{tab:supp_optim}
\begin{tabular*}{\columnwidth}{@{\extracolsep{\fill}}lcc@{}}
\toprule
\rowcolor{TableHeader} Parameter & Stage 1 & Stage 2 \\
\midrule
Scenes & 103k / 233k & 103k / 233k \\
Epochs & 25 & 40 \\
GPUs $\times$ batch / GPU & \multicolumn{2}{c}{$8\!\times\!6$} \\
Gradient accumulation & \multicolumn{2}{c}{1} \\
Optimizer & \multicolumn{2}{c}{AdamW} \\
Planner peak LR & \multicolumn{2}{c}{$1.73205\!\times\!10^{-4}$} \\
JEPA encoder LR & --- & $10^{-5}$ \\
JEPA predictor LR & --- & $10^{-4}$ \\
Encoder layer decay & --- & 0.9 \\
Weight decay: planner & \multicolumn{2}{c}{0.01} \\
Weight decay: JEPA & --- & 0.04 \\
Warm-up / schedule & \multicolumn{2}{c}{10\% / cosine} \\
Gradient norm clip & \multicolumn{2}{c}{1 per domain} \\
EMA momentum & --- & 0.99925 \\
Precision / seed & \multicolumn{2}{c}{BF16 / 2} \\
\bottomrule
\end{tabular*}
\end{minipage}\par\medskip

The image encoder uses four register tokens. The ego encoder projects the latest 11-D state to width 256. AR combines scene attention, a GRU, and bicycle integration. JEPA uses $16\!\times\!16$ patches and two-frame tubelets on $512\!\times\!256$ images. Each adapter projects features to width 256 and appends one encoded $4\!\times\!11$ ego-history token. DiT additionally attends to 64 scene tokens. The scorer combines trajectory embeddings, scene/ego attention, and gated video attention. We evaluate one scene per device with deterministic scene~noise.

AdamW uses $(\beta_1,\beta_2)=(0.9,0.999)$ and $\epsilon=10^{-8}$. The planner peak learning rate is $2\!\times\!10^{-4}\sqrt{48/64}$ with an initial warm-up multiplier of $10^{-6}$. JEPA biases and one-dimensional parameters have zero weight decay. Optimizer groups share the schedule multiplier. Resuming restores optimizer, scheduler, EMA, random states, and data position. Geometry, evaluator scoring, and detached observed-video encoding use FP32. Stage~2 epochs take about 2.5 hours on real data and 5.9 hours on mixed data.

\subsection{JEPA adaptation}
\label{app:video}
The online encoder, predictor, and EMA target start from the released Drive-JEPA checkpoint~\cite{wang2026drivejepa}. The checkpoint's complete driving-domain pretraining recipe is not publicly specified. We adapt its multiblock masking recipe to eight-frame rectangular clips. Recorded scenes provide four historical and ten future frames, and synthetic scenes provide four historical and eight future frames. We select eight contiguous, distinct frames at $0.5\pm0.1$\,s intervals, rotating windows with scene and epoch. Valid windows exist in 103,165 real and 128,445 synthetic training scenes. The remaining scenes contribute planning losses only.

The predictor has 12 layers, width 384, and 12 heads. Its two mask families use eight spatial blocks at scale 0.15 and two at scale 0.70, spanning time with aspect ratio $[0.75,1.5]$. Ten pretrained mask tokens are loaded, with groups zero and one assigned to these families. Front images are cropped by 28 pixels at the top and bottom, resized to $512\!\times\!256$, and ImageNet-normalized without horizontal flips. Visible tokens predict layer-normalized masked EMA features. This temporal masking does not impose strictly causal future prediction. A separate previous/current-frame pass supplies detached planning features. The predictor, EMA target, and future frames are absent at inference.

\subsection{Loss implementation}
\label{app:losses}
The objectives follow Section~\ref{sec:optimization}. Planning losses average valid candidates within each scene before batch reduction. Padding is masked, and labels, matching indices, and EMA targets are detached.

\paragraph{Trajectory targets.}
Expert anchoring uses mean summed absolute $x,y,\theta$ error over eight poses. The auxiliary long reference resamples ten recorded poses over 5\,s into the eight prediction slots with a cubic spline. For zero-based slot $k\in\{0,\ldots,7\}$, the source index is $k+(k+1)(k+2)/36$, corresponding to time $0.5[k+1+(k+1)(k+2)/36]$ seconds. This time-compressed auxiliary target ends at 5\,s; the planner's output timestamps remain 0.5--4\,s. Both expert coefficients are one. Missing synthetic long references contribute zero without changing batch normalization. Dense matching uses mean absolute position error plus 0.25 times wrapped heading error, with coefficient $0.25\min(1,e/2)$. Empty target sets contribute zero.

\paragraph{Replay.}
The 64 replay samples comprise 24 high-score ($J\geq0.95$), 24 medium-score ($0.8\leq J<0.95$), eight low-positive, and eight zero-score paths. Deficits are filled from unused nonzero candidates, and remaining padding is masked. Replay calibration has weight 0.25 and uses a separate attention context; ranking uses only native candidate pairs. The direction head receives BCE supervision with zero weight in the v1 aggregate.

\paragraph{Residuals and video.}
Compatible residual targets share one timestep per scene and one noise tensor per parent. Reconstruction error selects one pair for both $v$-prediction and geometric losses. Matched and unmatched parents are averaged separately, with unmatched weight 0.05 and zero contribution from empty groups. Fresh scorer labels come from target-free DDIM rollouts. The unit-weight video L1 objective averages feature dimensions, masked positions, both mask families, and globally valid clips in FP32. For $R$ DDP ranks with $n_r$ valid clips, rank $r$ scales its valid-clip mean by $Rn_r/\sum_q n_q$. Empty ranks contribute zero, and globally empty batches skip regression and EMA. EMA follows updates with video gradients.

\paragraph{Gradient paths.}
Generation updates image LoRA, scene/ego encoding, and AR. Residual learning updates shared scene features, DiT, and its adapter, with detached parents blocking direct AR gradients. Scoring updates shared scene/ego features, the scorer, and its adapter, with detached coordinates blocking direct DiT gradients. Detached video features isolate JEPA from planning losses. Only video regression updates its encoder and predictor. All objectives share one training loop, with separate clipping for planning and video gradients.

\subsection{Ablation and checkpoint protocols}
\label{app:ablation_protocol}
\paragraph{Stage~1.}
Real-data controls fix 103,263 scenes, seed~2, generic DINO initialization, 64 proposals, batch~48, and the schedule in Table~\ref{tab:supp_optim}. All-off disables only Dense, Replay, and Rank. Expert/long-expert losses, native-64 BCE, aggregate MSE, official soft labels, and data/replay forward paths are retained. All runs train from scratch for 25 epochs, or 53.8k updates. Mixed-data controls fix 232,672 scenes and differ only in Rank, training for 121.2k updates. Each row selects the highest PDMS among 25 complete benchmark evaluations. There is one training seed per configuration.

Oracle is the best official score among 64 candidates, and regret is Oracle minus selected score. Earlier real-data controls certify the maximum when the selected score is 100 and evaluate all candidates otherwise. Replay-off, mixed-data, and both single-target Stage~2 controls evaluate all 64 candidates per scene. Oracle and regret average over all 12,146~scenes.

\paragraph{Stage~2.}
Main DriveSTA and Scale results are selected from 19 and 17 complete evaluations, respectively. Real-data controls share seed~2, +Rank Stage~1 initialization, batch, data, a planned 40-epoch schedule (86.08k updates), and FP32 evaluation with 12 DDIM steps. The expert-target run is early-stopped. The screened one-target result is selected from 13 evaluations, and the adapted/frozen/removed video branches from 35/40/40 evaluations. As in Stage~1, each row reports the highest score among that run's complete evaluations.

Target controls retain AR dense supervision, residual parameterization, $10\mathcal L_v+20\mathcal L_\tau$, and scoring losses. The expert control assigns the reference to every parent before noise without proximity gates or a zero-residual fallback. Screened controls add both. Single-target assignment precedes noise, whereas four-target WTA follows reconstruction. Paired bootstrap intervals use 10,000 resamples of 44 driving logs with equally weighted scenes. Screened-minus-expert 95\% intervals are $[0.62,2.42]$ for PDMS, $[1.89,3.86]$ for Oracle, and $[0.89,1.71]$ for regret, conditional on the \mbox{selected checkpoints}.

\paragraph{Visual conditioning.}
The removed branch omits video features, ego-history tokens, and fusion layers. Frozen conditioning restores the entire branch. To isolate visual input, the fixed-pool diagnostic zeroes observed video only in the S2 scorer before adapter projection while retaining position and ego-history tokens. DiT receives video in both conditions. Model weights, scene/ego tokens, candidates, and official labels remain fixed. The diagnostic uses the 94.42 Stage~1 checkpoint and a Stage~2 checkpoint with native-pool PDMS 94.7240. Restoring video yields 95\% paired-bootstrap intervals of $[3.81,4.98]$ on initial proposals and $[2.57,3.78]$ on refined candidates, using the same 44-log resampling protocol.

\paragraph{Attention visualization.}
The attention visualizations in Figures~\ref{fig:teaser} and~\ref{fig:qualitative_collision} use the complete Scale S2 checkpoint from epoch 11 (step 58,176), with Navtest PDMS 95.5520. The cases were selected for visible traffic interactions and readable spatial patterns from the 41-scene visualization pool in Appendix~\ref{app:datasets}. Figure~\ref{fig:supp_jepa_attention} shows more localized scorer peaks and broader refiner attention around road users and the adjacent road in the selected scenes. Scorer maps use the selected candidate query and average all eight heads over 512 JEPA spatial keys ($16\!\times\!32$). Refiner maps use the corresponding mode and average all 12 DDIM steps, 12 layers, eight heads, and eight waypoints over 128 pooled JEPA keys ($8\!\times\!16$). Two observed frames form one temporal tubelet; maps are overlaid on the current frame and do not separate per-frame contributions. Each map uses its own first-to-99th-percentile display range with bilinear interpolation. Ego-history and DINO keys are excluded from the displayed spatial maps. Attention extraction leaves planner outputs unchanged. The comparison trajectories in Figures~\ref{fig:teaser} and~\ref{fig:qualitative_collision} retain their original evaluation records; supplementary attention-only BEV plans come from the same inference pass as the maps. These weights describe feature access, rather than the causal importance of individual image regions.

\subsection{Supplementary diagnostics}
\label{app:experiment_diagnostics}
\paragraph{Archive coverage.}
The real archive contains 317.3M paths, with 34.05\% scoring zero and 49.17\% scoring 100. Experts average 94.92, and 61.30\% of scenes contain a better path before screening. The synthetic census contains 629.2M paths, with 10.77\% zero and 43.00\% full scores against a reference mean of 79.66. Figure~\ref{fig:census} averages 128 real or 256 synthetic scenes per column. Reference curves summarize 24 equally populated groups in the same scene order.

Recovery fills 523 of 548 incomplete real pools. Retained scenes supply 26.44M high-score entries and 2.61M screened targets across 95,998 scenes. Targets average 99.66 per scene versus 95.27 for experts, and 55.45\% of scenes with targets contain an improvement. The mean best-target gain is 4.65, with a 95\% interval of $[4.38,4.94]$ from resampling 162 logs. Of 45,413 sub-95 experts, 45,253 lose weighted points only on progress. The 7,265 empty sets retain expert and score supervision.

\paragraph{Refinement and scoring.}
The paired diagnostic follows 8,192 parents from one adapted model across 128 scenes and 35 logs. Mean candidate PDMS rises from 75.16 to 81.05. Of 4,234 parents below 95, 1,660 cross the threshold. Of 3,958 already at or above it, 378 fall below, including 146 to zero. Oracle moves from 99.84 to 99.77, while the refined trajectory at the original best-parent index scores 89.49. In fixed-pool scorer transfer, Stage~1 scores 94.42 versus Stage~2's 93.86 on initial proposals, and Stage~2 scores 94.72 versus Stage~1's 93.53 on refined candidates.

\input{sections/12_appendix_inference_efficiency}

%% file: sections/12_appendix_inference_efficiency.tex
\subsection{Inference efficiency}
\label{app:efficiency}
Table~\ref{tab:inference_latency} times ChainFlow-VLA and the 232.7k \method{} checkpoints from GPU-resident inputs to the selected plan. Models run in turn on eight exclusive H20 GPUs, with one batch-one worker per GPU, 64 candidates, and identical shuffled scene-to-device assignments. Each of 12,146 scenes contributes one equally weighted measurement. Workers warm up for 30 calls, then two calls after each new 16-scene chunk. CUDA synchronization brackets wall-clock timing. Encoding, generation, refinement, scoring, and selection are included. Disk access, preprocessing, tokenization, transfers, and simulator scoring are excluded.

We use PyTorch 2.5.1/CUDA 12.4, with planning and JEPA in FP32 and TF32 disabled. ChainFlow-VLA uses BF16 InternVL, native eager attention, and no FlashAttention2. Every call recomputes features. ChainFlow-VLA and Stage~2 use 12 DDIM steps. Stage~1 omits video encoding and diffusion sampling.

%% file: sections/14_appendix_hugsim_breakdown.tex
\subsection{Detailed HUGSIM results}
\label{app:hugsim_detailed}
Table~\ref{tab:supp_hugsim_breakdown} gives the episode means for each execution metric, with PDMS and HD-Score computed per episode using \cref{eq:supp_hugsim}. In both training regimes, S2 improves average collision avoidance and time to collision but reduces comfort. For DriveSTA-Scale, these changes accompany a rise in mean HD-Score even as the Extreme subset drops from 30.3 to 27.3.

In Table~\ref{tab:hugsim}, \emph{Data} excludes pretraining. For the $^\dagger$ rows (DrivoR-Reg, LTF, and SimVTA), 103k is the size of the NAVSIM source split. SimVTA post-trains on the reconstructed subset~\cite{jing2026simvta}. Both Gigapixel entries count 335k self-play scenarios~\cite{rowe2026gigapixel}. The FPD-Drive results are preliminary~\cite{chen2026fpd}.

%% file: sections/14_appendix_hugsim_table.tex
\noindent\begin{minipage}{\textwidth}\centering\captionsetup{hypcap=false,skip=4pt}
\captionof{table}{\textbf{HUGSIM metric breakdown.} The four configurations and matched 345 episodes are those of Table~\ref{tab:hugsim}. All scores are scaled by 100; higher is better. E/M/H/X contain 49/126/86/84 episodes; Avg.\ is the scenario-weighted mean. Bold/underline mark best/second-best RC, HD-Score, and PDMS averages at displayed precision.}
\label{tab:supp_hugsim_breakdown}
\fontsize{8.5}{10}\selectfont
\setlength{\tabcolsep}{0pt}\renewcommand{\arraystretch}{1.25}
\begin{tabular}{@{}>{\raggedright\arraybackslash}p{94pt}|*{5}{>{\centering\arraybackslash}p{\dimexpr(\textwidth-94pt-3\arrayrulewidth)/15\relax}}|*{5}{>{\centering\arraybackslash}p{\dimexpr(\textwidth-94pt-3\arrayrulewidth)/15\relax}}|*{5}{>{\centering\arraybackslash}p{\dimexpr(\textwidth-94pt-3\arrayrulewidth)/15\relax}}@{}}
\toprule
\rowcolor{TableHeader}
 & \multicolumn{5}{c|}{RC $\uparrow$} & \multicolumn{5}{c|}{HD-Score $\uparrow$} & \multicolumn{5}{c}{PDMS $\uparrow$}\\
\rowcolor{TableHeader}
Method & E & M & H & X & Avg. & E & M & H & X & Avg. & E & M & H & X & Avg.\\
\midrule
\rowcolor{TableOurs}
\textbf{DriveSTA S1} & 83.1 & 56.9 & 37.1 & 41.2 & 51.9 & 73.8 & 41.0 & 20.7 & 25.7 & 36.9 & 83.6 & 55.2 & 30.4 & 36.6 & 48.5\\
\rowcolor{TableOurs}
\textbf{DriveSTA S2} & 84.3 & 68.0 & 35.2 & 39.5 & \underline{55.2} & 75.9 & 58.6 & 23.0 & 25.3 & \underline{44.0} & 85.9 & 76.9 & 43.6 & 40.7 & 61.1\\
\midrule
\rowcolor{TableOurs}
\textbf{DriveSTA-Scale S1} & 83.2 & 63.1 & 33.4 & 44.6 & 54.0 & 74.0 & 56.8 & 21.4 & 30.3 & \underline{44.0} & 90.3 & 87.9 & 38.6 & 42.7 & \underline{64.9}\\
\rowcolor{TableOurs}
\textbf{DriveSTA-Scale S2} & 84.5 & 67.9 & 37.8 & 40.6 & \textbf{56.1} & 75.2 & 61.5 & 27.6 & 27.3 & \textbf{46.7} & 88.3 & 90.3 & 52.1 & 44.7 & \textbf{69.4}\\
\bottomrule
\end{tabular}
\par\vspace{8pt}\nointerlineskip
\begin{tabular}{@{}>{\raggedright\arraybackslash}p{94pt}|*{5}{>{\centering\arraybackslash}p{\dimexpr(\textwidth-94pt-4\arrayrulewidth)/20\relax}}|*{5}{>{\centering\arraybackslash}p{\dimexpr(\textwidth-94pt-4\arrayrulewidth)/20\relax}}|*{5}{>{\centering\arraybackslash}p{\dimexpr(\textwidth-94pt-4\arrayrulewidth)/20\relax}}|*{5}{>{\centering\arraybackslash}p{\dimexpr(\textwidth-94pt-4\arrayrulewidth)/20\relax}}@{}}
\toprule
\rowcolor{TableHeader}
 & \multicolumn{5}{c|}{NC $\uparrow$} & \multicolumn{5}{c|}{DAC $\uparrow$} & \multicolumn{5}{c|}{TTC $\uparrow$} & \multicolumn{5}{c}{Comf. $\uparrow$}\\
\rowcolor{TableHeader}
Method & E & M & H & X & Avg. & E & M & H & X & Avg. & E & M & H & X & Avg. & E & M & H & X & Avg.\\
\midrule
\rowcolor{TableOurs}
\textbf{DriveSTA S1} & 89.6 & 62.8 & 38.9 & 42.7 & 55.7 & 97.3 & 97.9 & 95.9 & 96.4 & 96.9 & 87.3 & 55.6 & 31.0 & 42.5 & 50.8 & 91.4 & 96.4 & 95.6 & 92.5 & 94.5\\
\rowcolor{TableOurs}
\textbf{DriveSTA S2} & 92.1 & 84.5 & 53.5 & 48.9 & 69.2 & 97.7 & 98.3 & 96.4 & 96.1 & 97.2 & 90.4 & 79.4 & 45.6 & 47.2 & 64.7 & 88.4 & 89.6 & 88.2 & 87.2 & 88.5\\
\midrule
\rowcolor{TableOurs}
\textbf{DriveSTA-Scale S1} & 94.0 & 90.7 & 47.8 & 49.3 & 70.4 & 98.8 & 99.4 & 95.4 & 96.6 & 97.6 & 91.5 & 87.9 & 38.0 & 44.8 & 65.5 & 99.1 & 99.9 & 99.7 & 99.2 & 99.6\\
\rowcolor{TableOurs}
\textbf{DriveSTA-Scale S2} & 93.4 & 93.9 & 59.3 & 51.3 & 74.8 & 98.9 & 99.1 & 97.3 & 96.6 & 98.0 & 91.1 & 91.3 & 53.2 & 49.4 & 71.5 & 92.7 & 97.2 & 95.2 & 92.7 & 95.0\\
\bottomrule
\end{tabular}
\par\end{minipage}

%% file: sections/13_appendix_qualitative.tex
\section{Visualizations}
\label{app:qualitative}
\subsection{JEPA cross-attention}
\label{app:jepa_visualizations}
\begingroup\centering\captionsetup{hypcap=false,skip=4pt}
\begin{minipage}{\textwidth}\centering
\noindent\includegraphics[width=\linewidth]{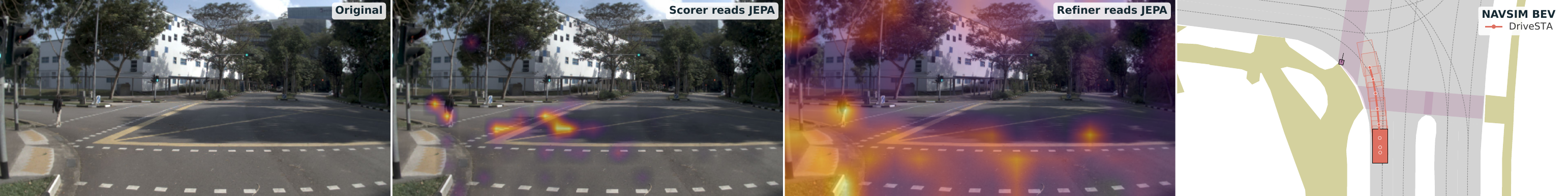}
\par\vspace{5pt}\nointerlineskip
\noindent\includegraphics[width=\linewidth]{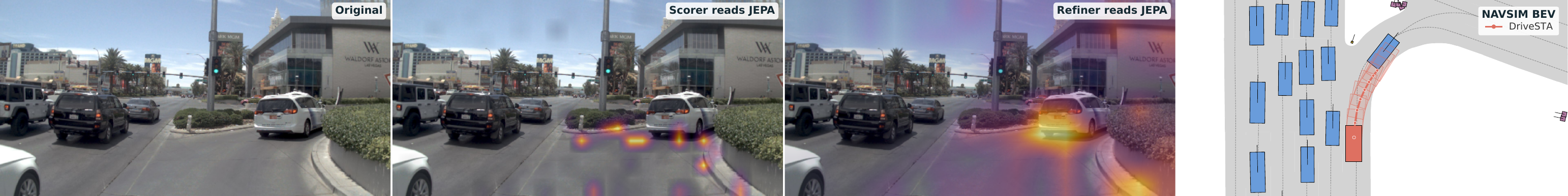}
\par\vspace{5pt}\nointerlineskip
\noindent\includegraphics[width=\linewidth]{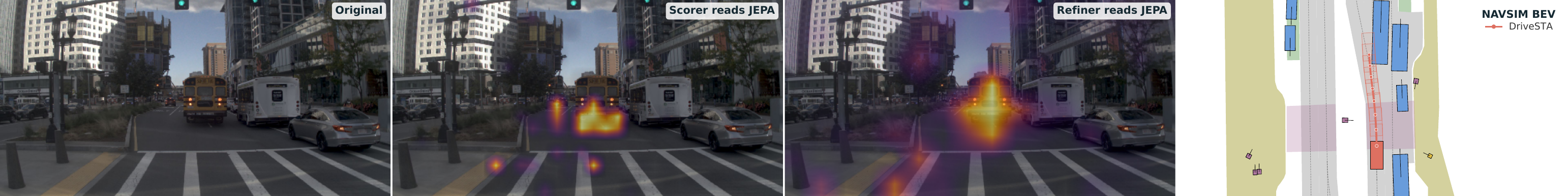}
\par\vspace{5pt}\nointerlineskip
\noindent\includegraphics[width=\linewidth]{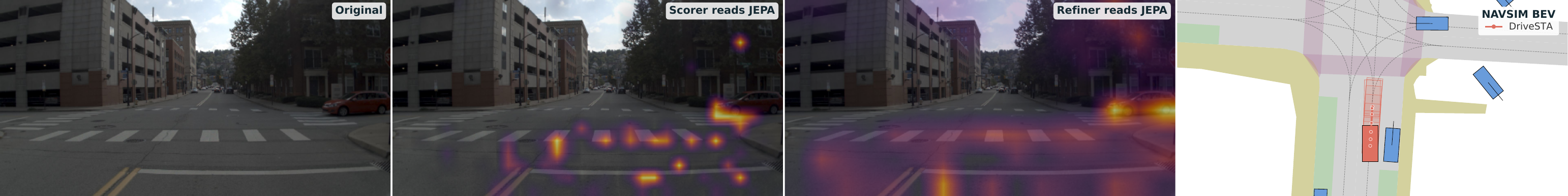}
\par\vspace{5pt}\nointerlineskip
\noindent\includegraphics[width=\linewidth]{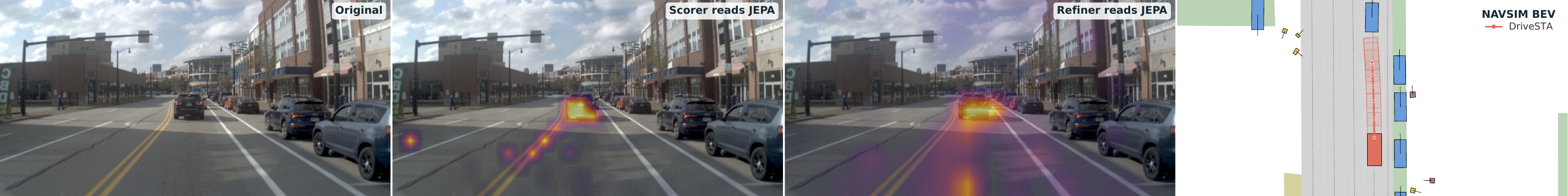}
\par\vspace{5pt}\nointerlineskip
\noindent\includegraphics[width=\linewidth]{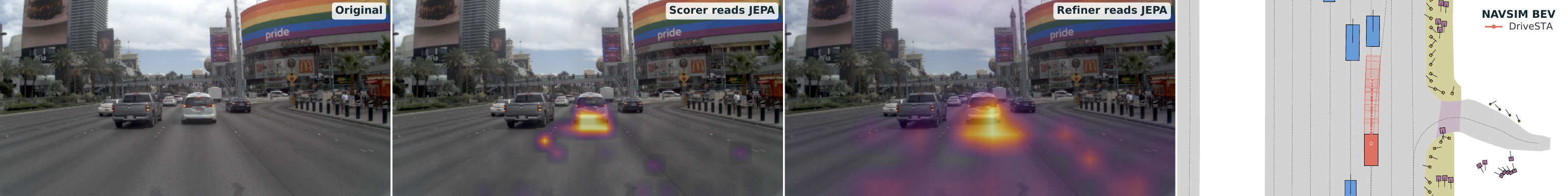}
\par\vspace{5pt}\nointerlineskip
\noindent\includegraphics[width=\linewidth]{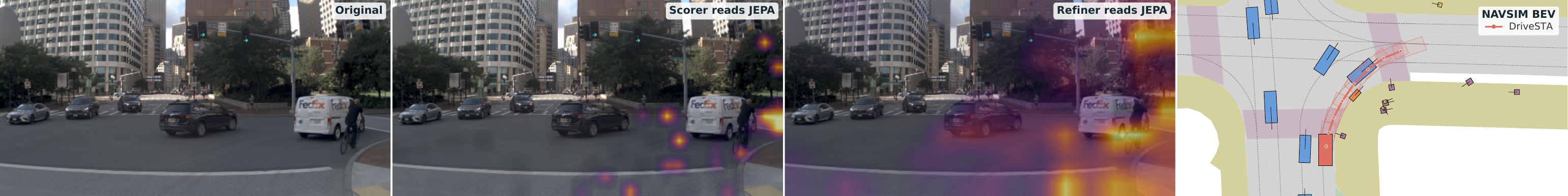}
\par\vspace{5pt}\nointerlineskip
\noindent\includegraphics[width=\linewidth]{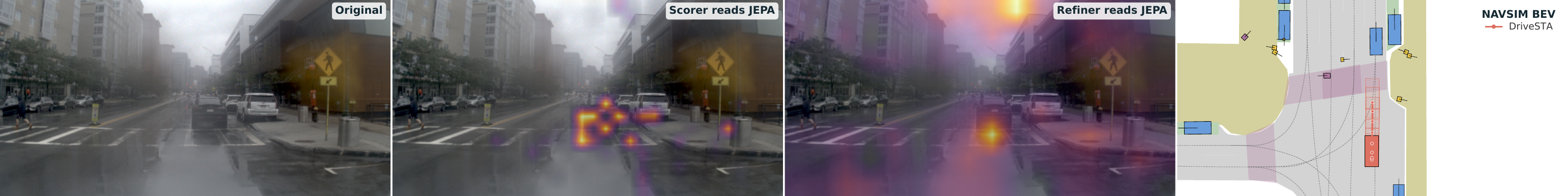}
\captionof{figure}{\textbf{Additional JEPA cross-attention examples.} The first four rows show the pedestrian, curved-lane, school-bus, and intersection examples. The last four rows show lead vehicles, a pedestrian at a junction, and backlit road traffic. Columns show the observed frame, scorer attention, refiner attention, and BEV plan. Refiner maps cover broader regions around visible traffic participants and adjacent road, whereas scorer peaks are more localized. Extraction and per-map normalization follow Appendix~\ref{app:ablation_protocol}. These selected cases illustrate spatial reading patterns rather than causal contributions.}
\label{fig:supp_jepa_attention}
\end{minipage}
\par\endgroup

\clearpage
\subsection{NAVSIM}
\newcommand{\suppNavNames}{%
\setlength{\unitlength}{\linewidth}%
\begin{picture}(1,0.025)
\put(0.1732,0.006){\makebox(0,0){\fontsize{8}{9}\selectfont Front view}}
\put(0.4245,0.006){\makebox(0,0){\fontsize{8}{9}\selectfont\bfseries DriveSTA}}
\put(0.5878,0.006){\makebox(0,0){\fontsize{8}{9}\selectfont FPD-Drive-Pro}}
\put(0.7512,0.006){\makebox(0,0){\fontsize{8}{9}\selectfont ChainFlow-VLA}}
\put(0.9146,0.006){\makebox(0,0){\fontsize{8}{9}\selectfont DrivoR-Scale}}
\end{picture}}
\begingroup\centering\captionsetup{hypcap=false,skip=4pt}
\begin{minipage}{\textwidth}\centering
\noindent\includegraphics[width=\linewidth]{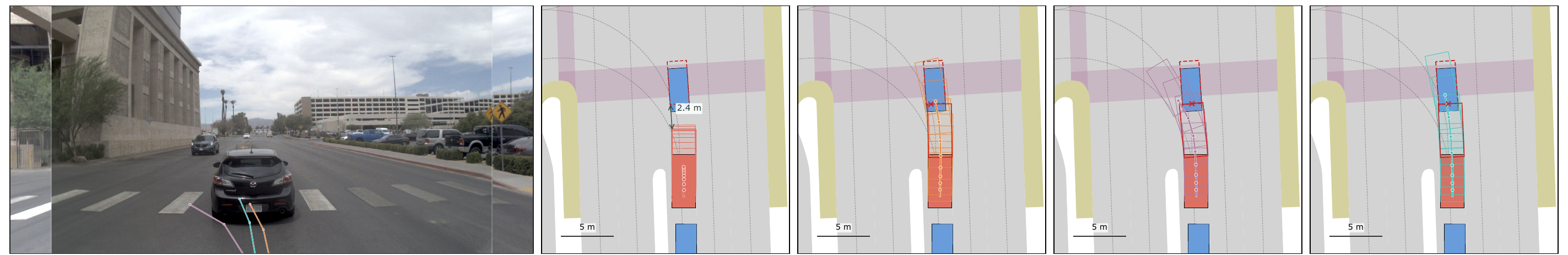}
\par\vspace{2pt}\nointerlineskip
\noindent\includegraphics[width=\linewidth]{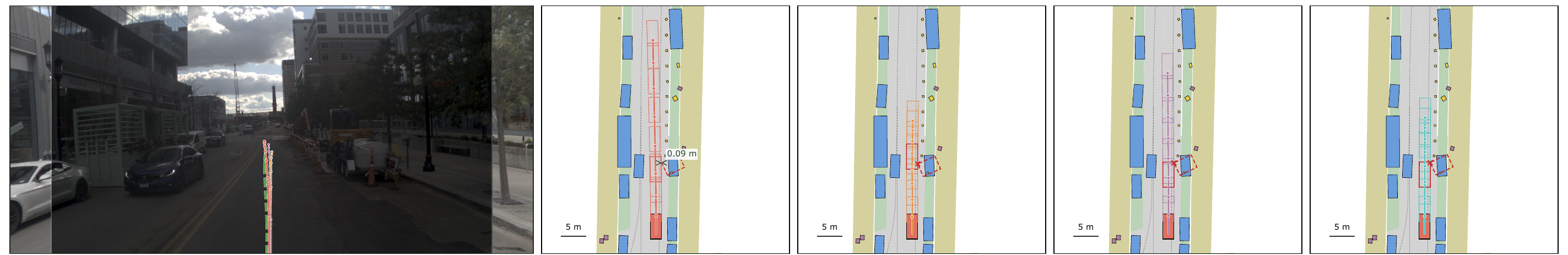}
\par\vspace{2pt}\nointerlineskip
\noindent\includegraphics[width=\linewidth]{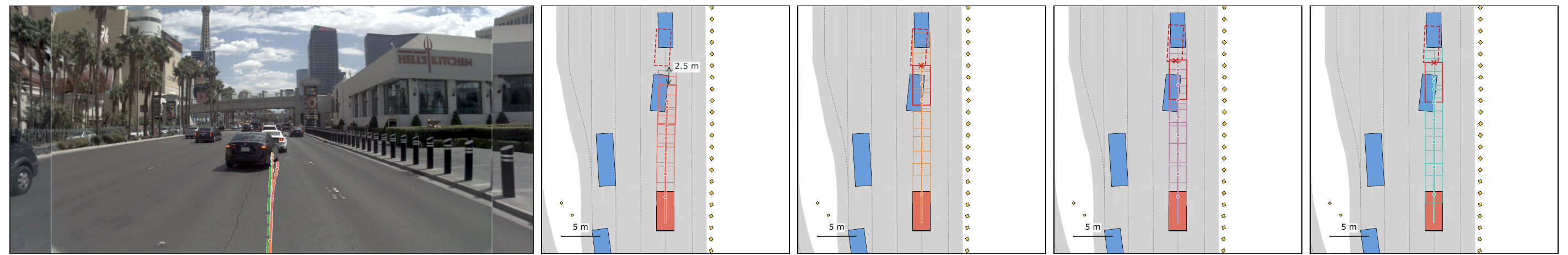}
\par\vspace{2pt}\nointerlineskip
\noindent\includegraphics[width=\linewidth]{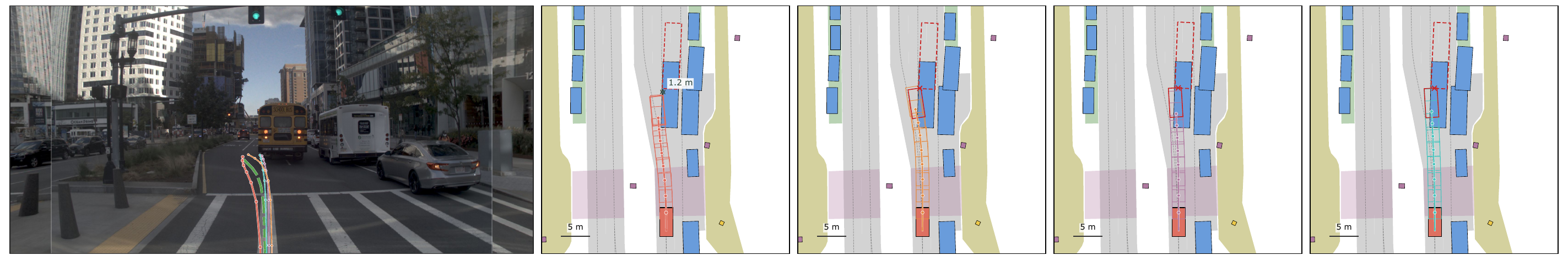}
\par\vspace{2pt}\nointerlineskip
\noindent\includegraphics[width=\linewidth]{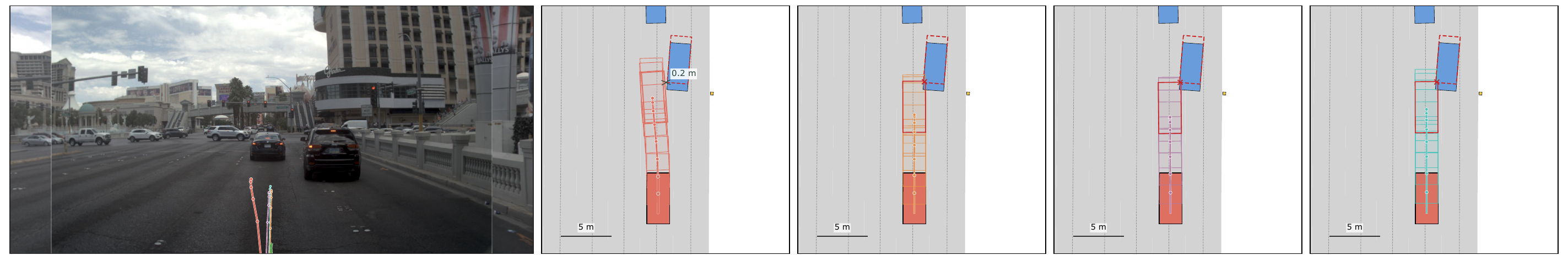}
\par\vspace{2pt}\nointerlineskip
\noindent\includegraphics[width=\linewidth]{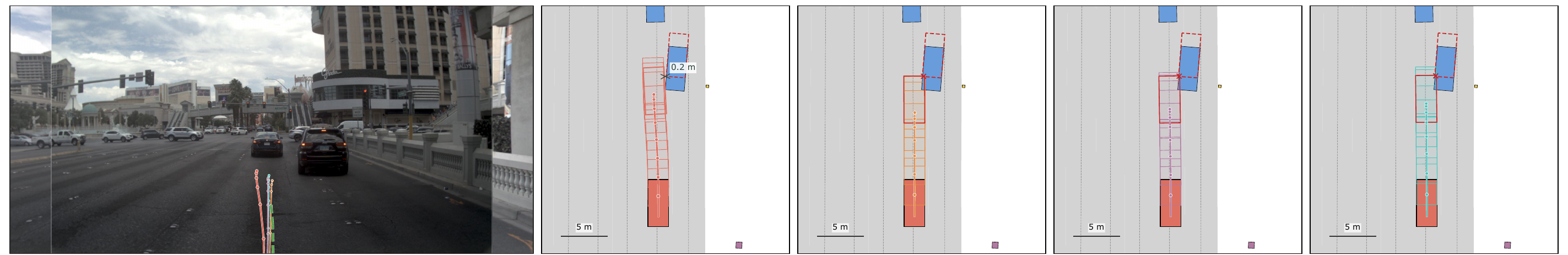}
\par\vspace{2pt}\nointerlineskip
\noindent\includegraphics[width=\linewidth]{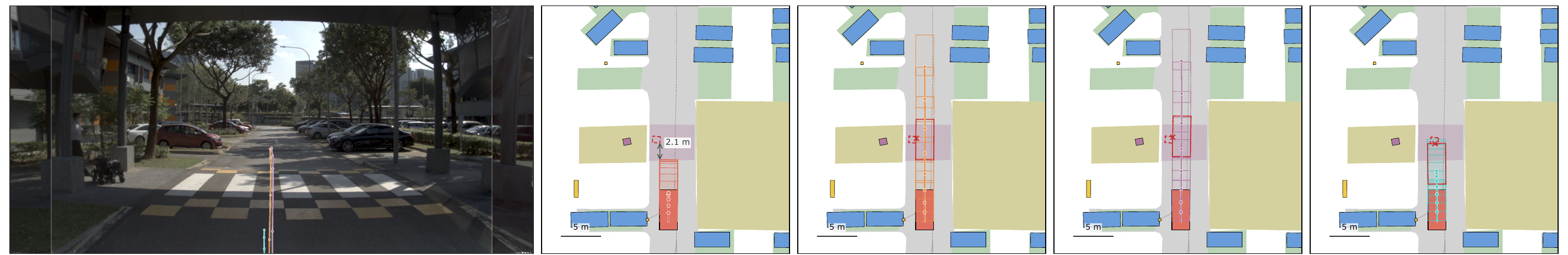}
\par\vspace{2pt}\nointerlineskip
\suppNavNames
\captionof{figure}{\textbf{Additional NAVSIM comparisons.} Each row shows one scene. Red crosses and dashed outlines mark contacts and contacted objects.}
\label{fig:supp_all_collide}
\end{minipage}
\par\endgroup

\clearpage
\begingroup\centering\captionsetup{hypcap=false,skip=4pt}
\begin{minipage}{\textwidth}\centering
\noindent\includegraphics[width=\linewidth]{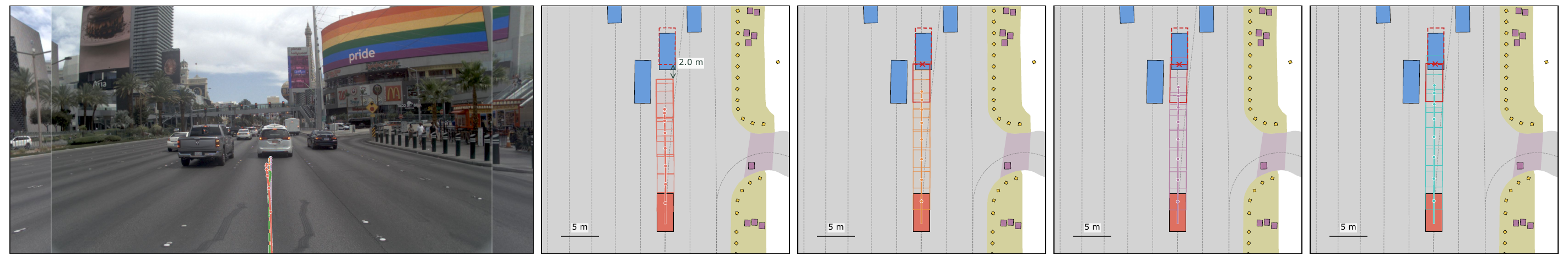}
\par\vspace{2pt}\nointerlineskip
\noindent\includegraphics[width=\linewidth]{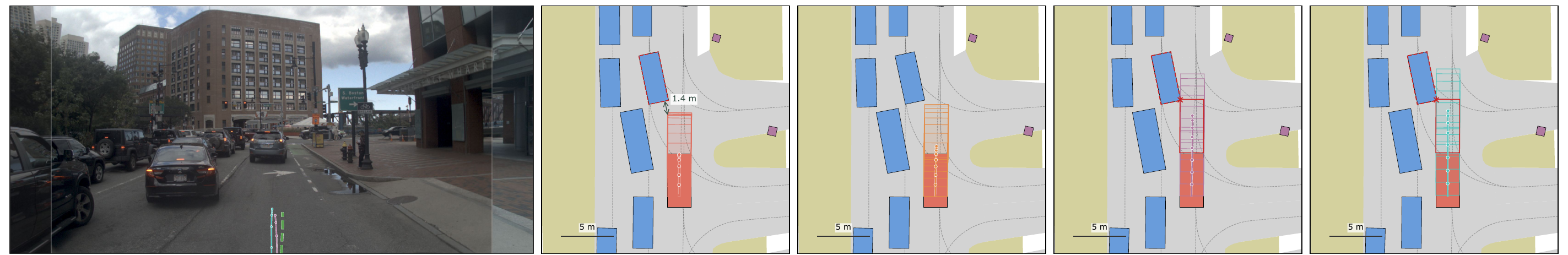}
\par\vspace{2pt}\nointerlineskip
\noindent\includegraphics[width=\linewidth]{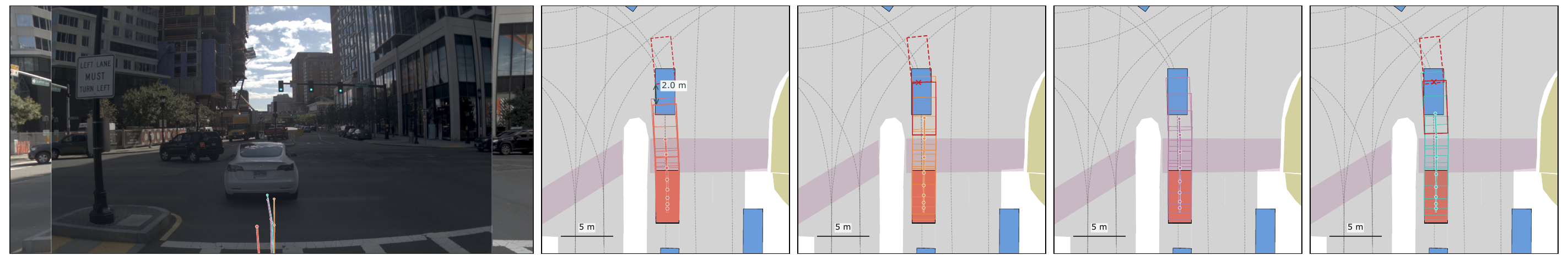}
\par\vspace{2pt}\nointerlineskip
\noindent\includegraphics[width=\linewidth]{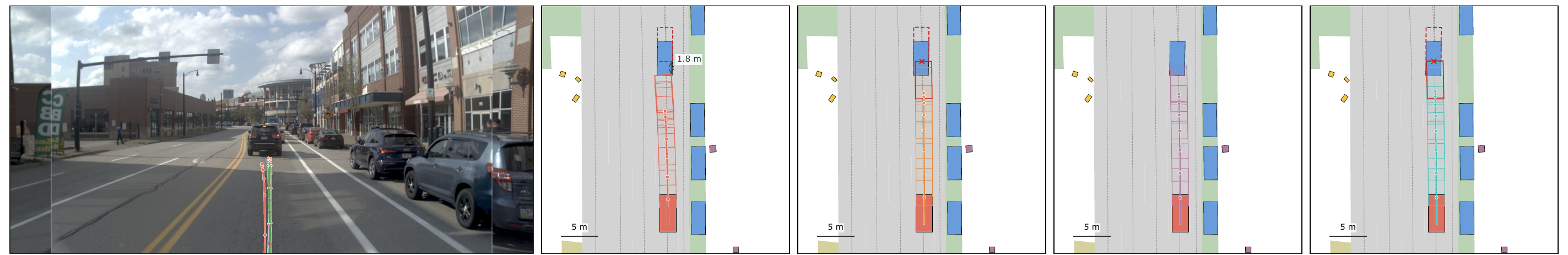}
\par\vspace{2pt}\nointerlineskip
\noindent\includegraphics[width=\linewidth]{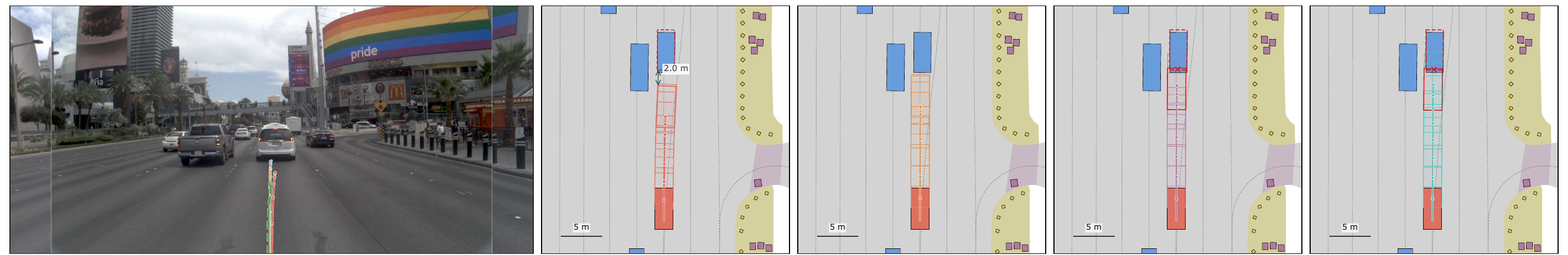}
\par\vspace{2pt}\nointerlineskip
\noindent\includegraphics[width=\linewidth]{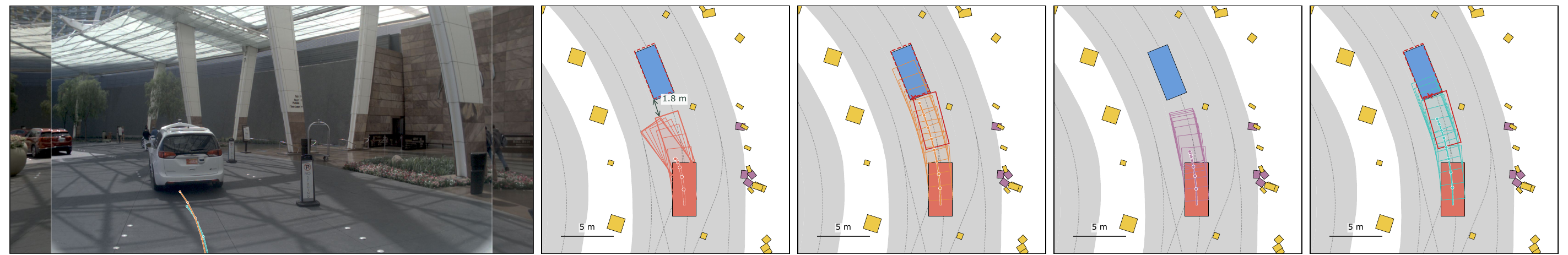}
\par\vspace{2pt}\nointerlineskip
\noindent\includegraphics[width=\linewidth]{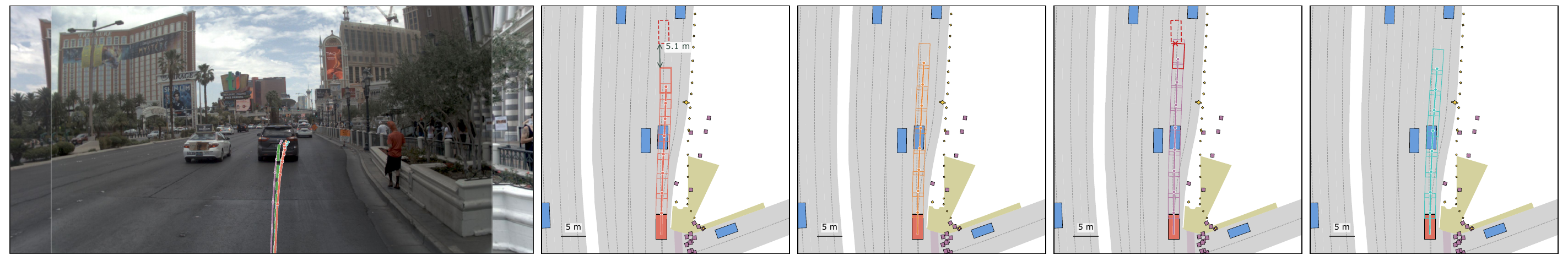}
\par\vspace{2pt}\nointerlineskip
\suppNavNames
\captionof{figure}{\textbf{Additional NAVSIM comparisons.} Each row shows one scene. Red crosses and dashed outlines mark contacts and contacted objects.}
\label{fig:supp_two_collide}
\end{minipage}
\par\endgroup

\clearpage
\subsection{HUGSIM}
\begingroup\centering\captionsetup{hypcap=false,skip=4pt}
\noindent\includegraphics[width=\textwidth]{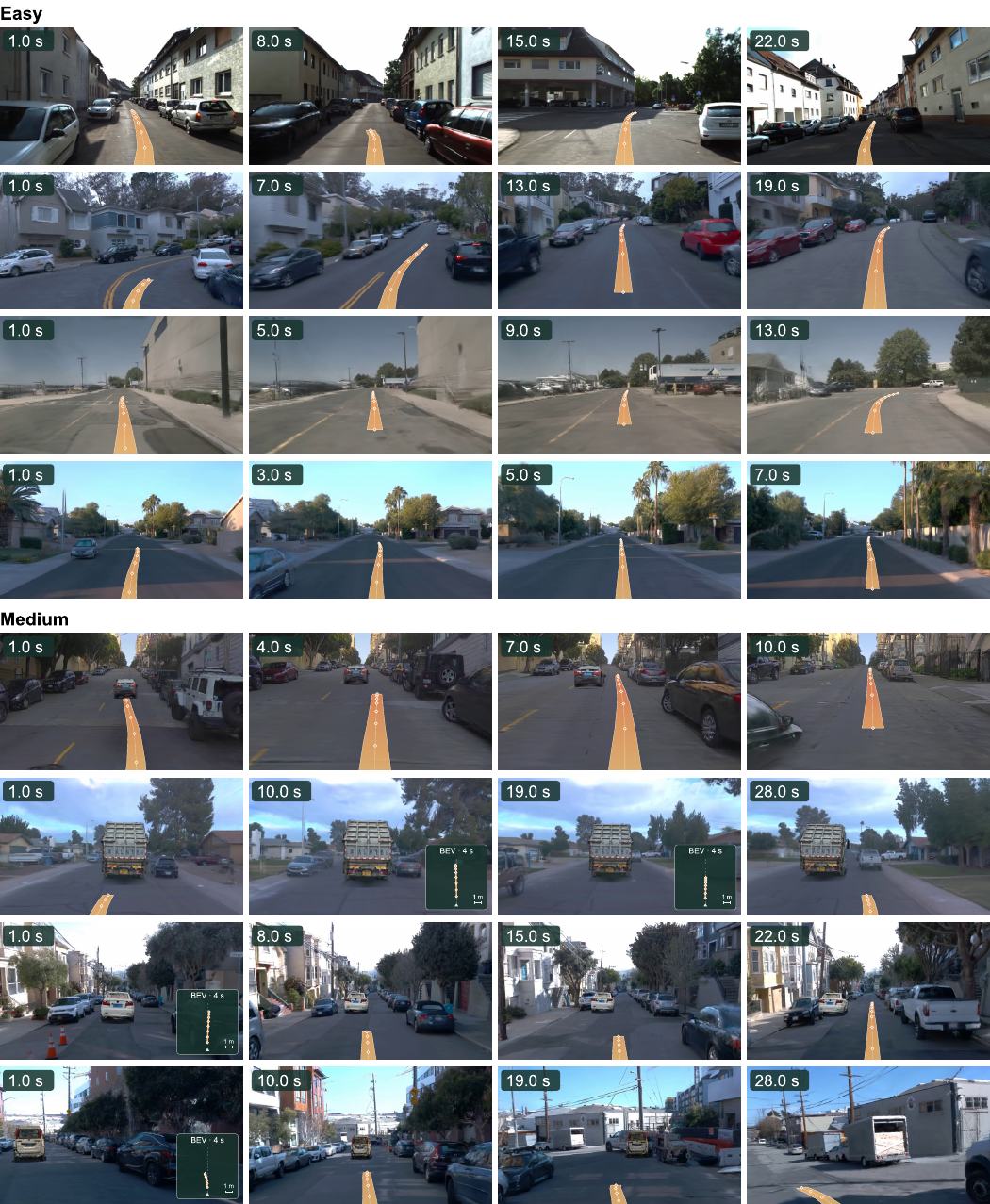}
\captionof{figure}{\textbf{Selected HUGSIM rollouts.} Orange shows DriveSTA-Scale S2 plans; insets use BEV. Labels give video time.}
\label{fig:supp_hugsim_1}
\endgroup

\clearpage
\begingroup\centering\captionsetup{hypcap=false,skip=4pt}
\noindent\includegraphics[width=\textwidth]{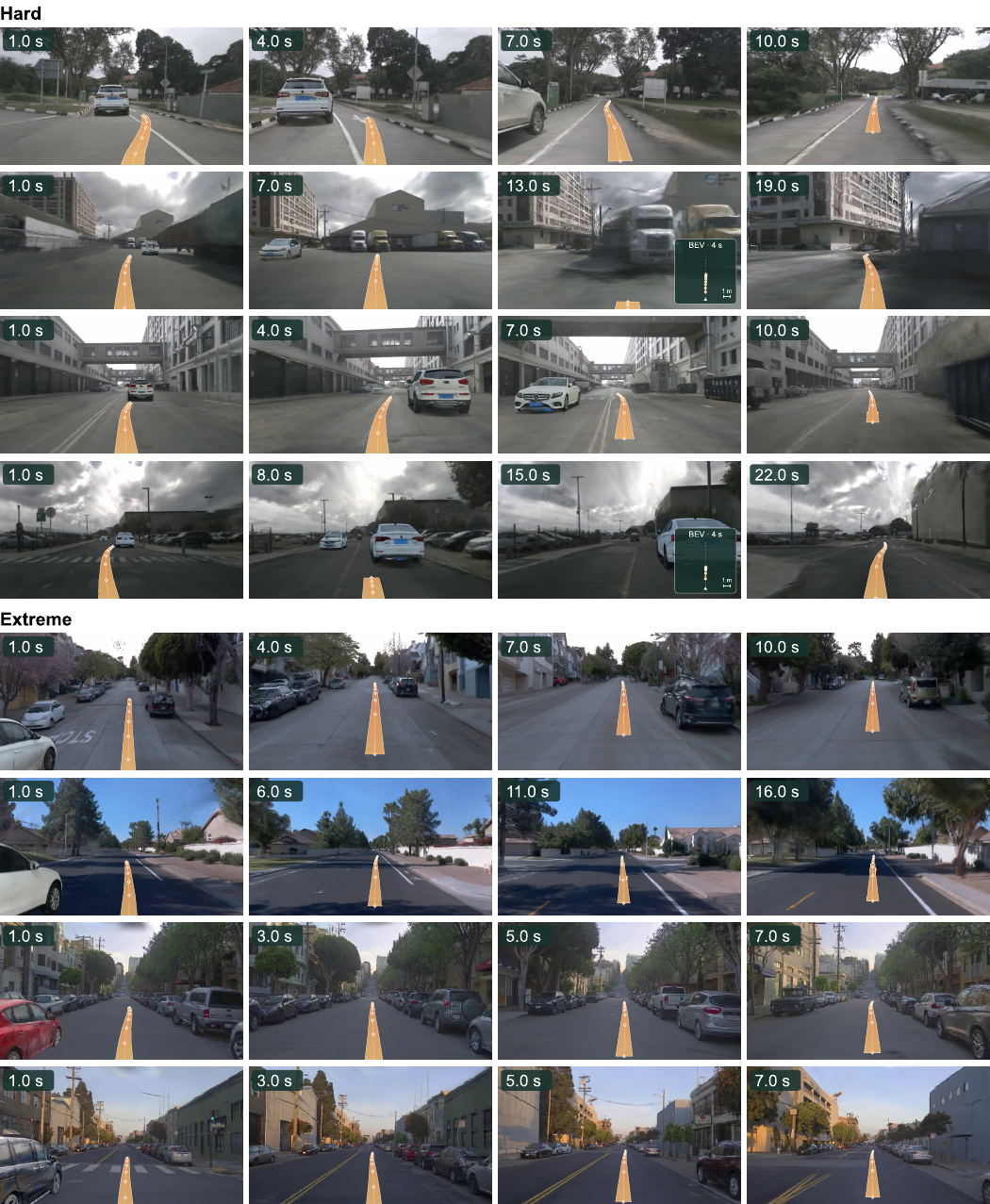}
\captionof{figure}{\textbf{Selected HUGSIM rollouts.} Orange shows DriveSTA-Scale S2 plans; insets use BEV. Labels give video time.}
\label{fig:supp_hugsim_2}
\endgroup